\documentclass[public]{boschreport}
\RequirePackage{needspace}

\title{\texorpdfstring{BoschDreams\textsuperscript{7V}}{BoschDreams-7V}: Controllable Video World Model for Long-Horizon Driving}
\brandtitle{BoschDreams}
\subtitle{Controllable Surround-View Video World Model for Long-Horizon Autonomous Driving}
\reporttagline{State-decoupled, memory-aware autoregressive diffusion streaming seven synchronized cameras in closed loop.}
\coverabstract{%
Closed-loop validation of autonomous driving requires surround-camera video streams that preserve 3D geometric coherence across overlapping views, remain visually stable over multi-minute rollouts, respond faithfully to counterfactual maneuvers, and execute at real-time sensor rates. We present \textbf{BoschDreams\textsuperscript{7V}}, a controllable seven-camera world model that decouples physical state evolution from neural sensor synthesis. An explicit 3D simulator advances traffic dynamics and projects HDMap vectors and oriented 3D bounding boxes into each calibrated camera. This lets a 28-block causal Diffusion Transformer focus on photorealistic rendering over a sparse 7-edge (14-directed-edge) physical rig adjacency graph. Training progresses through a four-stage curriculum—from a bidirectional surround teacher, block-causal student, and few-step self-forcing distillation to \emph{Antiphon Forcing}. This final stage combines a \emph{Layout-Aware Kernel Score} (LAK) with role-decoupled writer and denoiser low-rank adapters and decoder vector--Jacobian memory backpropagation to resolve inter-role gradient conflict. Paired with exact in-place rotary position rebasing and compiled FP8 kernels, BoschDreams\textsuperscript{7V} delivers two operating points on a 600-scene surround benchmark. In the \textbf{real-time 480p 1-step} tier, it streams all seven cameras synchronously at \textbf{10.85\,FPS} per camera (exceeding the 10\,Hz camera clock) with \textbf{80.19} seven-view FVD, \textbf{7.02\,m} ego-trajectory ADE, and \textbf{119.53} late-window FVD ($-68.0\%$ vs.\ Stage~III). In the \textbf{high-fidelity 720p 4-step} tier, it achieves \textbf{84.35} seven-view FVD (\textbf{99.67} late-window FVD), \textbf{4.90\,m} ADE, and \textbf{380.50} cross-view LoFTR matches. Adapted as a stochastic flow-bridge fixer, it improves degraded 3D Gaussian Splatting renders by a mean of \textbf{$+$1.26\,dB} PSNR (up to $+$2.12\,dB) while reducing FID by \textbf{29.6\%} (up to 38.0\%), FVD by \textbf{34.7\%} (up to 45.7\%), and BEV feature distance (FBD) by \textbf{17.5\%} (up to 25.6\%).%
}
\reporttype{Technical Report}
\docversion{1.0}
\reportdate{November 2026}
\author{Bosch XC World Model Team}
\entity{Robert Bosch GmbH}
\copyrightyear{2026}
\projectpage{https://jiangchaokang.github.io/boschdreams-7v-report/}

\providecommand{\degree}{\ensuremath{^\circ}}
\counterwithout{equation}{chapter}
\newcommand{\sg}{\operatorname{sg}}
\newcommand{\E}{\mathbb{E}}
\newcommand{\R}{\mathbb{R}}
\newcommand{\Normal}{\mathcal{N}}

\newcommand{\yes}{\textcolor{boschgreen}{\ensuremath{\checkmark}}}
\newcommand{\no}{\textcolor{boschgray}{\ensuremath{\times}}}
\newcommand{\opt}{\textcolor{boschblue}{Opt.}}
\newcommand{\na}{---}

\renewcommand{\hd}[1]{\textbf{#1}}
\newcommand{\best}[1]{\textbf{#1}}
\newcommand{\secbest}[1]{\underline{#1}}
\newcommand{\subrow}[1]{{\scriptsize\color{boschgray}#1}}
\newcommand{\tablenote}[1]{\par\vspace{2.5pt}{\footnotesize\color{boschgray}#1\par}}
\renewcommand{\arraystretch}{1.16}

\begin{document}

\maketitle

\raggedbottom
\setcounter{tocdepth}{2}
\tableofcontents
\clearpage

\needspace{5\baselineskip}
\section{Introduction}\label{sec:intro}

Validating an autonomous driving stack against rare, safety-critical events—such as an occluded scooter cutting across an urban crosswalk or an aggressive highway merge—cannot rely on passive fleet logs alone. Hazardous interactions occur too rarely, and recorded pixels cannot react when the ego planner alters its trajectory. A closed-loop generative sensor simulator must therefore satisfy four coupled demands. Its seven surround cameras must preserve 3D geometric coherence as vehicles cross overlapping fields of view; its autoregressive stream must resist visual drift over multi-minute rollouts; its observations must obey explicit road-map and agent controls; and its inference engine must meet the 10\,Hz camera clock while supporting automated scenario orchestration.

Existing simulation paradigms leave a clear structural gap (\cref{tab:capability}). Traditional graphics engines~\citep{dosovitskiy2017carla} and 3D Gaussian Splatting (3DGS) reconstructions~\citep{kerbl2023gaussians,chen2025omnire} preserve explicit 3D geometry, yet graphics assets suffer from a lighting and sensor-noise domain gap, while 3DGS renders tear into detached floaters under large counterfactual viewpoint shifts. Conversely, forward-view driving world models~\citep{hu2023gaia,wang2023drivedreamer,gao2024vista,horizondrive} lack 360\degree\ situational awareness, and bidirectional multi-view diffusion models~\citep{gao2025magicdrive2,ren2025cosmosdrivedreams,nvidia2025cosmos25} require tens of full-clip denoising evaluations rather than low-latency causal streaming. Even recent autoregressive multi-view generators~\citep{zheng2026xworld,xworld16,xiaomiwm1,xiaomiwm2,cheng2026m4world,nvidia2026omnidreams,nvidia2026cosmosinteractive} remain constrained in camera count, real-time throughput, unbounded positional horizon, or automated fleet-ticket scenario orchestration.

Our core insight is twofold: decoupling explicit 3D world-state authority from neural appearance synthesis enforces cross-view geometry at the input, while decoupling memory writing from few-step denoising inside the generator eliminates long-horizon rollout drift. An upstream kinematic simulator advances agent states and deterministically projects high-definition map (HDMap) polylines and oriented 3D bounding boxes into all seven calibrated cameras (\cref{fig:projection}), relieving pixel attention from implicitly solving multi-body kinematics. Within the causal generator, assigning separate parameters to key--value (KV) memory commits and incoming-chunk denoising prevents the two roles from cancelling each other's gradients during long-horizon post-training.

Guided by this principle, we build \textbf{BoschDreams\textsuperscript{7V}} around a sparse seven-camera Diffusion Transformer (DiT; \cref{fig:network}), a four-stage curriculum culminating in \emph{Antiphon Forcing} (\cref{sec:curriculum}), a real-time streaming runtime with exact rotary position rebasing (\cref{sec:runtime}), and an autonomous AI-agent loop that converts fleet failure tickets into edited closed-loop test worlds (\cref{sec:closedloop}). Across 600 held-out test scenes (\cref{sec:experiments}), BoschDreams\textsuperscript{7V} provides two complementary operating points. The real-time 480p 1-step configuration streams all seven cameras synchronously at 10.85 frames per second (FPS) per camera ($2.14\times$ faster than the 4-step eager baseline, surpassing the 10\,Hz sensor clock), achieving 80.19 seven-view FVD (74.93 on front wide), 7.02\,m ego-trajectory ADE, and 119.53 late-window FVD ($-68.0\%$ vs.\ Stage~III). The high-fidelity 720p 4-step configuration reaches 84.35 seven-view FVD (99.67 late-window FVD), 4.90\,m ADE, and 380.50 cross-view LoFTR matches, while continuous surround rollouts remain photometrically stable past 209\,s (\cref{tab:longhorizon,fig:longhorizon}).

Our primary contributions are:
\begin{enumerate}
  \item \textbf{State-decoupled sparse seven-view architecture:} A layout-grounded 28-block causal DiT that encodes projected HDMap and 3D box layouts via the causal VAE and fuses them with noisy video latents before patchification without increasing token sequence length, restricts cross-view attention to a 7-undirected-edge (14-directed-edge) physical rig adjacency graph ($66.7\%$ fewer cross-camera pairs than dense multi-view attention), and shares weights across all seven cameras via adaptive layer normalization (\cref{sec:method}; \cref{tab:main,fig:perview}).
  \item \textbf{Four-stage curriculum and Antiphon Forcing:} A progressive training pipeline from bidirectional surround teacher to causal student, self-forcing distillation, and memory-decoupled Stage~IV post-training combining our \emph{Layout-Aware Kernel Score} (LAK), role-specific writer/denoiser adapters, and decoder vector--Jacobian memory backpropagation (\cref{sec:curriculum}; \cref{fig:telemetry,fig:stage_progression,tab:ablation}).
  \item \textbf{Real-time streaming, 3DGS restoration, and AI-agent simulation:} A zero-sync compiled FP8 streaming engine with exact rotary position rebasing reaching 10.85 per-camera FPS (\cref{fig:latency,fig:rebase,tab:supp_efficiency}), a paired flow-matching fixer that improves 3DGS restoration by a mean of $+1.26$\,dB PSNR (up to $+2.12$\,dB) while reducing FID by $29.6\%$ (up to $38.0\%$), FVD by $34.7\%$ (up to $45.7\%$), and BEV feature distance (FBD) by $17.5\%$ (up to $25.6\%$) (\cref{sec:fixer,tab:fixer}), and an end-to-end closed-loop AI-agent scenario orchestration pipeline (\cref{sec:closedloop}).
\end{enumerate}

\begin{table}[tbp]
  \centering
  \caption{\textbf{System-level capability comparison of representative driving world models.} Entries follow each work's published technical report or official release specification (\no: not supported; \texttt{n/r}: not reported; $<10$: below the $10$\,Hz synchronous per-camera RGB threshold in the cited configuration).}
  \label{tab:capability}
  \scriptsize
  \setlength{\tabcolsep}{2.4pt}
  \begin{tabular*}{\linewidth}{@{\extracolsep{\fill}}lccccccccc@{}}
    \toprule
    \hd{Method} & \hd{Views} & \hd{Duration} & \hd{Control} & \hd{First Frame} & \hd{NFE / Chunk} & \hd{$\ge$10\,FPS/Cam} & \hd{Hardware} & \hd{AI Agent} \\
    \midrule
    DriveDreamer~\citep{wang2023drivedreamer} & 1 & Seconds & L, E, I, T & \opt & n/r & $<10$ & GPU & \no \\
    MagicDrive-V2~\citep{gao2025magicdrive2} & 6 & $\le$20\,s & L, E, C, T & \opt & n/r & $<10$ & GPU & \no \\
    GAIA-2~\citep{russell2025gaia2} & 5 & Seconds$^\star$ & L, E, C, T & \opt & 50--100 & $<10$ & H100 & \no \\
    Cosmos-Transfer 2.5 (Auto)~\citep{nvidia2025cosmos25} & 7 & Seconds$^\star$ & L, E, T, V & \opt & 70 (35$\times$2) & $<10$ & GPU & \no \\
    X-World 1.0~\citep{zheng2026xworld} & 7 & 24\,s & L, E, C, T & \opt & 4 (+KV) & n/r & n/r & \no \\
    X-World 1.6~\citep{xworld16} & 7 & $>$30\,s & L, E, C, T, G & \opt & 5 (4+1W) & $<10^\ddagger$ & PPU M810E & \no \\
    Xiaomi Auto WM 1.0~\citep{xiaomiwm1} & 1 / 3 & $\le$1\,min & L, E, C, T, G & \yes & 4 & $<10^\ddagger$ & H20 & \no \\
    Xiaomi Auto WM 2.0~\citep{xiaomiwm2} & 1--7 & Minutes & L, E, C, T, G & \opt & 2 & n/r$^\ddagger$ & H20 & \no \\
    HorizonDrive~\citep{horizondrive} & 1 & $\le$1\,min & L, E, T & \yes & 4 & $<10^\ddagger$ & RTX 5090 & \no \\
    OmniDreams~\citep{nvidia2026omnidreams} & 1 / 4 & Minutes & L, E, T & \opt & 3 (2+1W) & \yes & GB300 & \no \\
    \midrule
    \rowcolor{boschblue!7}
    \textbf{BoschDreams\textsuperscript{7V} (ours)} & \textbf{7} & \textbf{Multi-min.$^\dagger$} & L, E, T & \opt & \textbf{2} (1+1W) & \yes & GPU / DCU & \yes \\
    \bottomrule
  \end{tabular*}
  \tablenote{Control: L layout/boxes/road semantics, E ego action/trajectory, C camera geometry, T text, I image, G 3DGS/point-cloud prior, V visual control maps. First frame: \opt\ supports both image-conditioned and zero-RGB cold-start modes. NFE / Chunk: DiT passes per chunk ($+1\text{W}$ denotes a KV-memory commit pass). $^\star$Supports multi-window autoregressive extension. $^\ddagger$Below 10\,FPS decoded RGB per camera or reported at latent rate. $^\dagger$Demonstrated continuously to $209$\,s with constant KV memory (\cref{eq:rebase}).}
\end{table}

\needspace{5\baselineskip}
\section{Related Work}\label{sec:related}

\subsection{Controllable Driving World Models}\label{sec:rel_world}
Generative driving world models synthesize sensory observations conditioned on ego actions and scene structure. Forward-view formulations—including discrete-token autoregression in GAIA-1~\citep{hu2023gaia} and DrivingWorld~\citep{hu2024drivingworld}, conditional diffusion in DriveDreamer~\citep{wang2023drivedreamer}, Vista~\citep{gao2024vista}, and DriveDreamer-2~\citep{zhao2025drivedreamer2}, and causal diffusion in Epona~\citep{zhang2025epona} and HorizonDrive~\citep{horizondrive}—established trajectory- and layout-guided synthesis, yet single-camera streams lack 360\degree\ situational awareness.

Building on multi-sensor benchmarks such as nuScenes~\citep{caesar2020nuscenes}, multi-camera world models extend conditioning across full vehicle rigs. MagicDrive~\citep{gao2023magicdrive}, DriveWM~\citep{wang2023drivewm}, MagicDrive-V2~\citep{gao2025magicdrive2}, and UniMLVG~\citep{chen2025unimlvg} generate six-camera videos from projected 3D boxes, BEV/HD maps, and ego actions. GAIA-2~\citep{russell2025gaia2} and GAIA-3~\citep{wayve2025gaia3} apply space--time factorized latent diffusion to multi-camera fleet rigs conditioned on ego kinematics and 3D cuboids, while the Waymo World Model~\citep{waymo2026worldmodel} extends foundation world modeling to joint camera--LiDAR simulation. Cosmos-Drive-Dreams~\citep{ren2025cosmosdrivedreams}, Cosmos-Predict~2.5, and Cosmos-Transfer~2.5~\citep{nvidia2025cosmos25} scale scenario control to seven-camera rigs via multi-step bidirectional sampling (70 guided NFE per 93-frame clip). More recently, streaming multi-view systems—X-World~1.0/1.6~\citep{zheng2026xworld,xworld16}, Xiaomi Auto World Model~1.0/2.0~\citep{xiaomiwm1,xiaomiwm2}, M\textsuperscript{4}World~\citep{cheng2026m4world}, NVIDIA OmniDreams~\citep{nvidia2026omnidreams}, and Interactive Cosmos~\citep{nvidia2026cosmosinteractive}—distill bidirectional teachers into causal few-step generators with rolling KV caches. Whereas OmniDreams~\citep{nvidia2026omnidreams} demonstrates real-time two-step streaming on one- and four-camera deployments in AlpaSim, BoschDreams\textsuperscript{7V} targets synchronized seven-camera 360\degree\ streaming via a sparse 7-edge (14-directed-edge) physical rig adjacency graph, role-decoupled post-training, and exact rotary key rebasing.

\subsection{Autoregressive Diffusion Distillation and Long-Horizon Memory}\label{sec:rel_distill}
Standard diffusion and rectified-flow models~\citep{lipman2022flowmatching,liu2022rectifiedflow,esser2024sd3} require tens of sequential evaluations per sample, motivating few-step interactive world models such as LingBot-World~\citep{team2026lingbot}, HY-WorldPlay~\citep{sun2025worldplay}, and FastVideo-distilled generators~\citep{zhang2025fastvideo,zhang2025vsa}. Distribution Matching Distillation (DMD)~\citep{yin2024dmd} compresses sampling into one to four steps via score matching with an auxiliary paired regression term, and DMD2~\citep{yin2024dmd2} removes that paired term through two-time-scale critic updates. To adapt bidirectional video models to causal streaming, Diffusion Forcing~\citep{chen2024diffusionforcing} trains under independent per-frame noise levels, CausVid~\citep{yin2024causvid} introduces block-causal asymmetric DMD distillation, and Self Forcing~\citep{huang2025selfforcing} closes the train--inference exposure gap by unrolling the few-step student on its own generated KV cache.

Subsequent methods further improve causal initialization and rollout horizon. Causal Forcing~\citep{zhu2026causalforcing} and Causal Forcing++~\citep{zhao2026causalforcingpp} use autoregressive ODE initialization to enforce frame-level injectivity; Self-Forcing++~\citep{cui2025sfpp} and LongLive~\citep{yang2026longlive} supervise segments from extended student rollouts; Connected Self Forcing~\citep{zhang2026csf} replays cross-chunk gradients through clean KV writes; and Memory Forcing~\citep{huang2025memoryforcing} and Prospective Forcing~\citep{shen2026prospectiveforcing} investigate spatial memory and multi-step look-ahead. Most closely related to our Stage~IV, SGF+~\citep{su2026sgfplus} identifies negative gradient alignment between context writing and denoising in single-view video generation and assigns separate parameters to each role. Antiphon Forcing builds on this role-decoupling principle and extends it to layout-conditioned seven-camera driving simulation through three additions: (i) lightweight rank-32 dual-role LoRA adapters on a frozen surround backbone; (ii) a paired-reference Layout-Aware Kernel Score (LAK)~\citep{gneiting2007scoring,gretton2012mmd} with a codec-calibrated HDMap probe; and (iii) truncated causal-VAE decoder-VJP backpropagation into cached surround KV states.

\subsection{Closed-Loop Simulation and Neural Scene Restoration}\label{sec:rel_sim}
Closed-loop validation traditionally relies either on game-engine simulators such as CARLA~\citep{dosovitskiy2017carla}, whose artist-authored assets diverge from real sensor statistics, or on 3D/4D neural and Gaussian Splatting reconstructions~\citep{kerbl2023gaussians,yan2024streetgaussians,NEURIPS2024_aa9508b2,chen2025omnire}, alongside complementary streaming vector-graph world models~\citep{jiang2026vectorworld} and pseudo auto-labeling of 3D dynamic scene flow~\citep{jiang20243dsflabelling}. While dynamic 3DGS methods such as OmniRe~\citep{chen2025omnire} reconstruct rigid and non-rigid actors at high frame rates, large off-trajectory maneuvers still expose sparse-splat floaters, missing background regions, and seam artifacts. Recent works mitigate novel-view degradation either by feeding generative priors back into 4D scene optimization as in DriveDreamer4D~\citep{zhao2025drivedreamer4d}, or by training single- and few-step diffusion restoration enhancers—including Difix3D+~\citep{wu2025difix}, DiffusionHarmonizer~\citep{zhang2026diffusionharmonizer}, ArtiFixer~\citep{delutio2026artifixer}, and the paired reconstruction Diffusion Fixer in OmniDreams~\citep{nvidia2026omnidreams}—related mathematically to Schrödinger bridges and stochastic interpolants~\citep{liu2023i2sb,albergo2023interpolants}. BoschDreams\textsuperscript{7V} connects explicit kinematic traffic simulation~\citep{kong2015bicycle,treiber2000idm,kazemkhani2024gpudrive} and neural rendering along two paths: as a layout-conditioned seven-camera streaming generator coupled to interactive policies or a ticket-driven scenario agent, and as a seven-camera stochastic source-perturbed flow-bridge fixer ($\sigma_{\text{src}}=0.3$) evaluated across five canonical 3DGS degradation families.

\needspace{5\baselineskip}
\section{Model Architecture}\label{sec:method}

\begin{figure}[tbp]
  \centering
  \includegraphics[width=\linewidth,height=0.50\textheight,keepaspectratio]{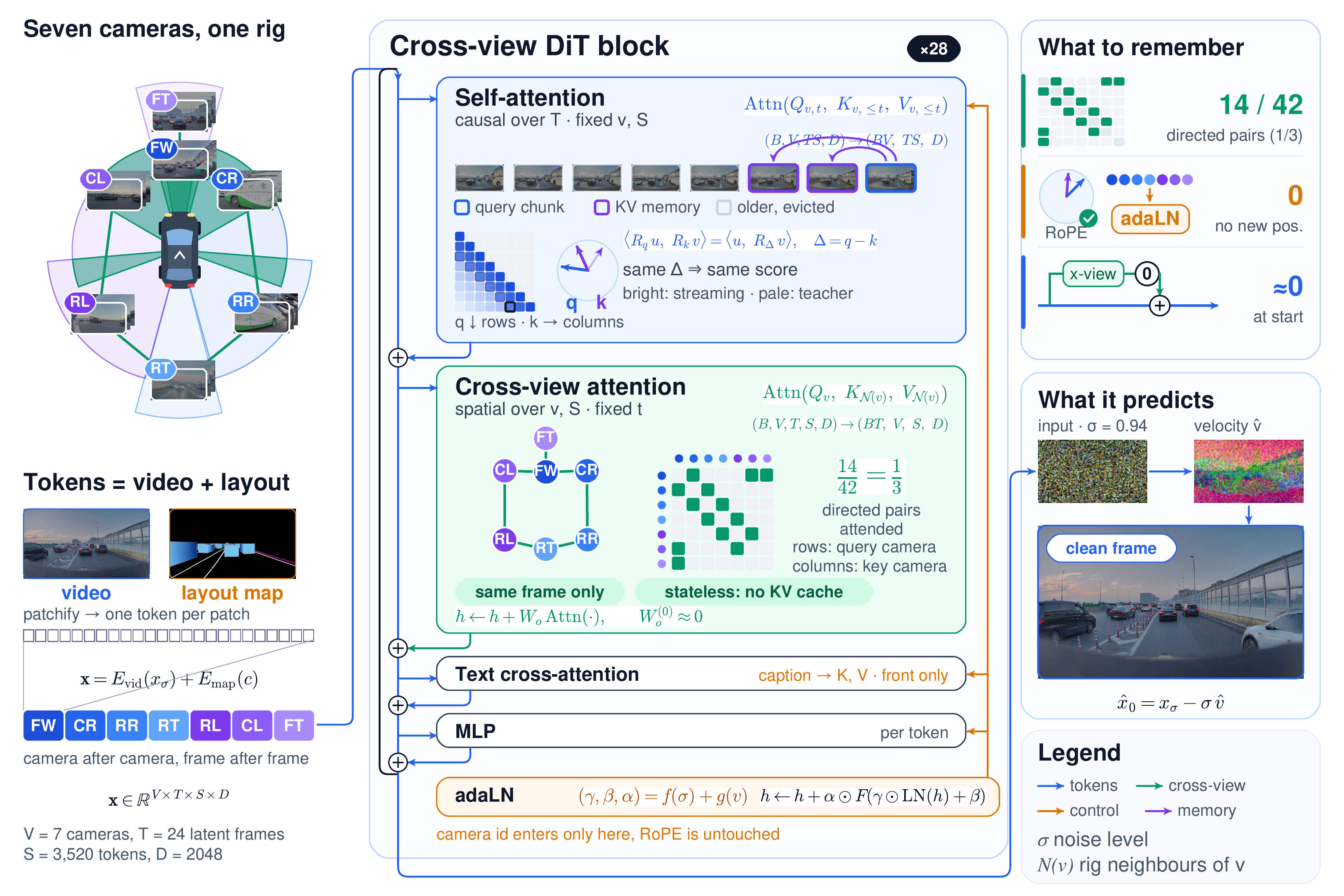}
  \caption{\textbf{BoschDreams\textsuperscript{7V} 28-block causal DiT architecture.} Video latents $z_{t,v}$ and causal-VAE-encoded HDMap/3D-box layout latents $z^{\text{layout}}_{t,v}$ across seven cameras are patchified on the same spatial grid and summed in embedding space (\cref{eq:fusion}). Each block alternates intra-camera causal temporal attention (\cref{eq:time}), sparse cross-view spatial attention over physical rig neighbors $\mathcal{S}(v)$ (\cref{eq:xview}), and text cross-attention, with flow time $\sigma$ and camera index $v$ injected via \texttt{adaLN} (\cref{eq:adaln}).}
  \label{fig:network}
\end{figure}

\subsection{State--Appearance Separation and Multi-Camera Formulation}\label{sec:formulation}
\begin{figure}[tbp]
  \centering
  \includegraphics[width=0.92\linewidth,height=0.48\textheight,keepaspectratio]{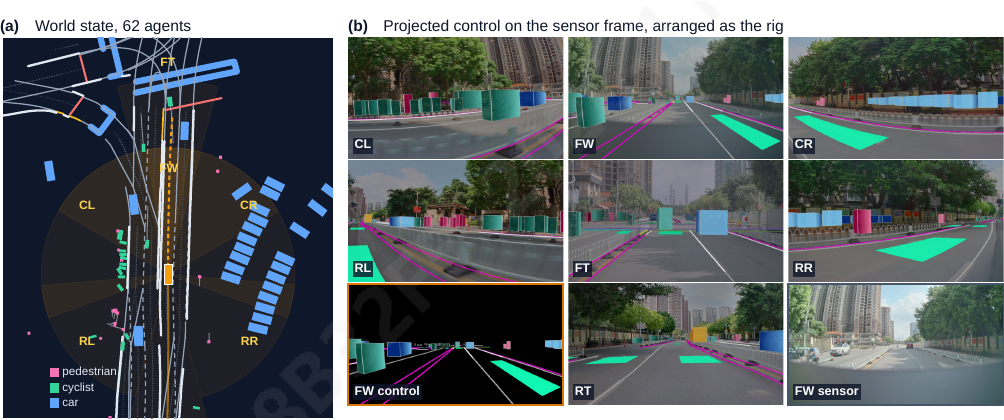}
  \caption{\textbf{Multi-camera 3D layout projection from a unified world state.} (a) Bird's-eye view of a held-out urban intersection showing HDMap polylines, 62 tracked dynamic agents with 1.5\,s motion trails, ego trajectory (solid past, dashed future), and seven camera frusta. (b) Perspective projections composited over recorded camera images, with lower corners decomposing \texttt{FW} into its layout control map and raw sensor frame.}
  \label{fig:projection}
\end{figure}

Monolithic video world models that infer agent dynamics and camera pixels within a single neural network force visual capacity to implicitly solve multi-body kinematics, leading to vanishing vehicles, distorted lane topology, and cross-camera disagreement. BoschDreams\textsuperscript{7V} resolves this by separating physical state authority from neural appearance synthesis (\cref{fig:network,fig:projection}). Let $\mathcal{C}=\{1,\dots,C\}$ index the $C=7$ rigidly mounted cameras of the ego vehicle (calibrated intrinsics and extrinsics in \cref{tab:rig}; abbreviations and mathematical notation in \cref{tab:abbrev,tab:notation}). At each timestamp, an upstream state simulator—driven either by logged LiDAR-fused 3D annotations, reactive traffic policies, or an AI scenario agent—maintains the unified 3D scene state and deterministically projects static HDMap vectors (lane dividers, road edges, crosswalks, stop lines) and dynamic oriented 3D bounding boxes into camera-aligned semantic layout frames $M_t=\{m_{t,v}\}_{v=1}^C\in\R^{C\times H\times W\times C_m}$.

Conditioned on the projected layout stream $M_{1:T}$, a scene-level text caption $y$ (encoded via frozen Cosmos-Reason1-7B~\citep{nvidia2025cosmosreason1}), and an optional initial sensor frame, the generator synthesizes synchronized seven-camera video $X_{1:T}\in\R^{T\times C\times H\times W\times 3}$ autoregressively over contiguous chunks $X^{(k)}$ of $L=2$ latent frames ($8$ display frames, or $0.8$\,s at $10$\,Hz):
\begin{equation}
  p_\theta(X_{1:T}\mid M_{1:T},y)=\prod_{k=1}^{K}p_\theta\big(X^{(k)}\mid X^{(<k)},M^{(k)},y\big).
  \label{eq:factor}
\end{equation}
Because all seven layout maps $m_{t,1:C}$ are perspective projections of a single 3D world state, inter-camera object position, scale, and heading consistency are enforced directly at the input interface. The diffusion transformer (\cref{fig:network}) devotes its representational budget to rendering photorealistic lighting, surface material, weather, and coherent temporal memory.

\begin{figure}[tbp]
  \centering
  \includegraphics[width=\linewidth,height=0.50\textheight,keepaspectratio]{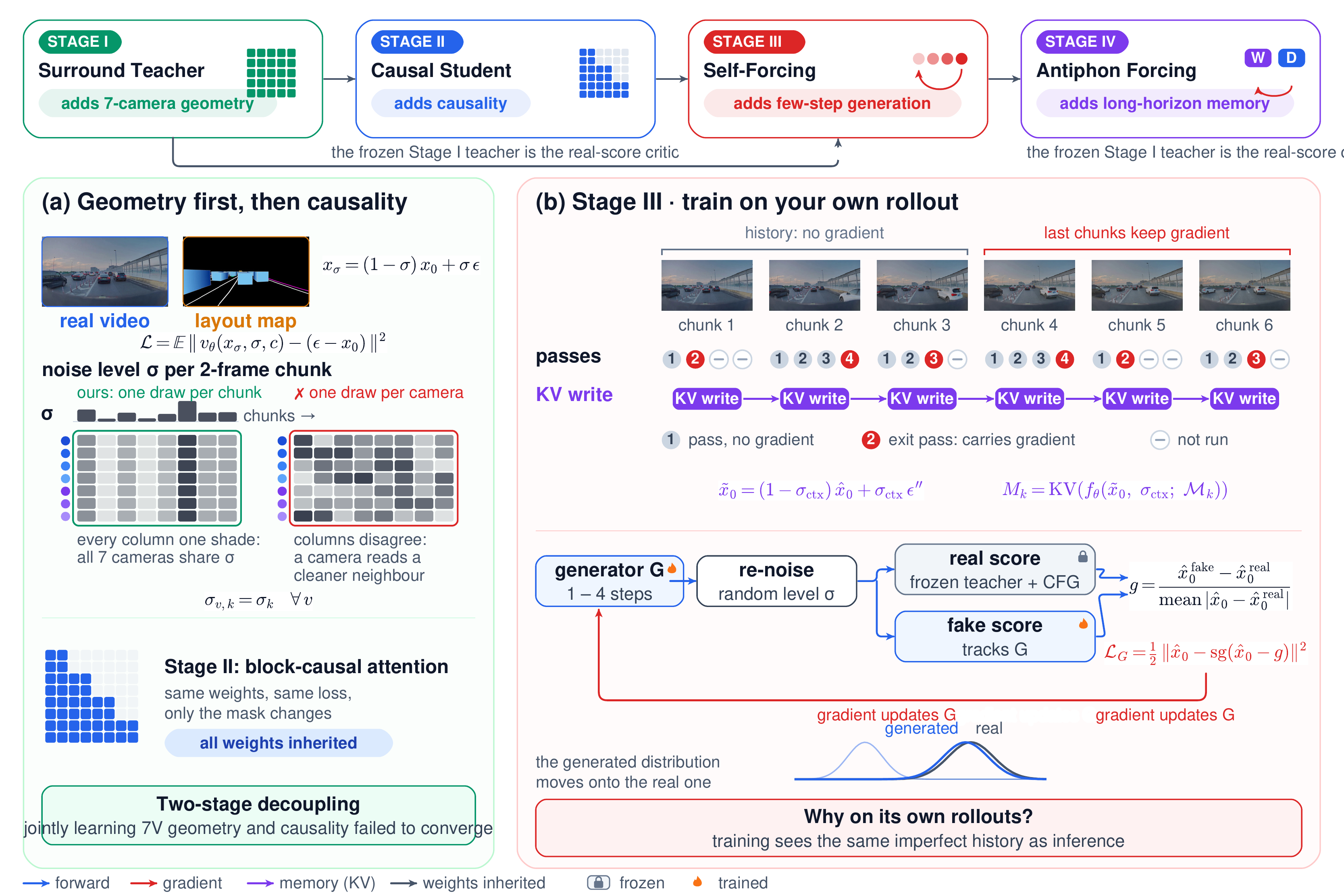}
  \caption{\textbf{Curriculum Stages I--III.} (a) Stage~I trains a bidirectional 7-view teacher with shared per-chunk noise $\sigma$ across all seven cameras; Stage~II inherits all weights and applies a 2-latent block-causal mask (\cref{eq:mask}). (b) Stage~III unrolls the few-step student $G_\phi$ on its own KV memory writes and supervises exit chunks via rollout DMD (\cref{eq:dmd}) against the frozen Stage~I teacher.}
  \label{fig:curriculum}
\end{figure}

\subsection{Additive Layout Tokenization and Factorized Rig Attention}\label{sec:attention}
Processing raw seven-camera video directly is prohibitive at streaming rates. A causal 3D Variational Autoencoder (VAE; Wan~2.1~\citep{wan2025}) compresses each camera stream by $8\times$ spatially and $4\times$ temporally ($T_{\text{lat}}=1+(T_{\text{pix}}-1)/4$, so latent frame $0$ depends solely on initial RGB frame $0$). Each 16-channel latent $z_{t,v}\in\R^{16\times h\times w}$ is partitioned into $2\times2$ spatial patches, yielding $N_s=(h/2)(w/2)$ tokens of dimension $D=2048$ per frame ($1{,}560$ tokens/camera at 480p; $3{,}520$ at 720p; \cref{tab:config}). Concatenating layout tokens along the sequence axis would double sequence length and quadruple self-attention FLOPs. Instead, the same causal VAE encodes each layout frame $m_{t,v}$ into a manifold-aligned 16-channel latent $z^{\text{layout}}_{t,v}\in\R^{16\times h\times w}$, which is channel-concatenated with the noisy video latent—equivalently summing linear $2\times2$ patch projections $\mathcal{F}_{\text{patch}}$ and $\mathcal{F}_{\text{map}}$:
\begin{equation}
  u^{0}_{v,t,i}=\mathcal{F}_{\text{patch}}(z_{t,v,i})+\mathcal{F}_{\text{map}}\big(z^{\text{layout}}_{t,v,i}\big)\in\R^{D},
  \label{eq:fusion}
\end{equation}
where $i\in\{1,\dots,N_s\}$ indexes spatial patches. This additive patch fusion injects pixel-aligned road and bounding-box geometry at zero extra sequence length.

Across 28 identical DiT blocks~\citep{peebles2022dit}, spatiotemporal attention is factorized into two specialized operators (\cref{fig:network}) rather than joint 3D multi-view attention:
\begin{enumerate}
  \item \textbf{Intra-camera causal temporal attention:} Within each camera $v$, query tokens at latent frame $q$ attend exclusively to same-camera keys at frames $k\le q$ across the active two-frame chunk and retained KV cache. Spatial patch coordinates are encoded via standard 2D spatial RoPE, while temporal positions are modulated by 1D temporal Rotary Position Embeddings (RoPE)~\citep{su2021roformer} $\mathbf{R}(\cdot)$ (rotation and re-basing in \cref{eq:rebase} act along the temporal axis only):
  \begin{equation}
    \mathbf{A}_{\text{time}}(Q_v,K_v,V_v)=\operatorname{Softmax}\!\left(\frac{(\mathbf{R}(q)Q_{v,q})(\mathbf{R}(k)K_{v,k})^{\top}}{\sqrt{d_k}}+\mathbf{M}_{\text{causal}}(q,k)\right)V_{v,k},
    \label{eq:time}
  \end{equation}
  where head dimension $d_k=128$ and $\mathbf{R}(q)^{\top}\mathbf{R}(k)=\mathbf{R}(k-q)$ ensures temporal translation invariance.
  \item \textbf{Sparse physical rig-adjacency cross-view attention:} Across cameras at simultaneous latent frame $t$, dense attention over all $C(C-1)=42$ directed cross-camera pairs expends most of its compute on distant or non-adjacent viewpoints (e.g., front telephoto to rear telephoto). We instead restrict cross-view attention to a 7-undirected-edge (14-directed-edge) \emph{physical rig adjacency graph} $\mathcal{E}$ connecting each wide/side camera to its immediate ring neighbors and each coaxial/rear telephoto camera to its flanking neighbors (\mbox{FW--FT}, \mbox{FW--CL}, \mbox{FW--CR}, \mbox{CL--RL}, \mbox{CR--RR}, \mbox{RL--RT}, and \mbox{RR--RT}; $14/42 = 33.3\%$ of off-diagonal directed pairs, or $21/49 = 42.9\%$ of all key--query pairs including self-attention). Letting $\mathcal{S}(v)=\{v\}\cup\mathcal{N}(v)$ with $\mathcal{N}(v)=\{u:(v,u)\in\mathcal{E}\}$, camera $v$ attends only to keys and values in $\mathcal{S}(v)$:
  \begin{equation}
    \mathbf{u}^{\text{xview}}_{v,t}=\operatorname{Softmax}\!\left(\frac{Q_{v,t}\big[K_{u,t}\big]_{u\in\mathcal{S}(v)}^{\top}}{\sqrt{d_k}}\right)\big[V_{u,t}\big]_{u\in\mathcal{S}(v)}.
    \label{eq:xview}
  \end{equation}
\end{enumerate}

To share a single set of backbone parameters across heterogeneous optics ($30^\circ$ telephoto to $120^\circ$ wide-angle) without binding spatial patch tokens to fixed camera embeddings, flow noise level $\sigma\in[0,1]$ and camera index $v\in\mathcal{C}$ enter exclusively through adaptive layer normalization (\texttt{adaLN}):
\begin{equation}
  \operatorname{adaLN}(h,\sigma,v)=\big(1+\gamma(\sigma,v)\big)\odot\frac{h-\mu_h}{s_h}+\beta(\sigma,v),
  \label{eq:adaln}
\end{equation}
where $\mu_h,s_h$ denote channel mean and standard deviation, and modulation vectors $(\gamma,\beta,\alpha)$ (with residual gate $\alpha$) are predicted once per camera frame by a shared MLP.

\subsection{Reconciling Real-Data Distribution Fidelity with HDMap Controllability}\label{sec:objective}
A central challenge in controllable world modeling is preventing structural layout conditioning from degrading visual realism—or, conversely, preventing distribution-matching post-training from eroding adherence to HDMap lanes and 3D agent boxes. BoschDreams\textsuperscript{7V} couples three mechanisms across its training stages:

First, during Stages~I and II (\cref{sec:stages123}), the velocity network $v_\theta$ is trained via Rectified Flow Matching~\citep{liu2022rectifiedflow} on recorded video latents $z_0$ paired with 3D-projected layouts $M$ from LiDAR/SLAM annotations. Defining linear interpolants $z_\sigma=(1-\sigma)z_0+\sigma\epsilon$ for $\epsilon\sim\Normal(0,\mathbf{I})$ and $\sigma\sim\mathcal{U}(0,1)$, the network minimizes:
\begin{equation}
  \mathcal{L}_{\text{flow}}(\theta)=\E_{\sigma,z_0,\epsilon,M,y}\Big[\big\|v_\theta(z_\sigma,\sigma,M,y)-(\epsilon-z_0)\big\|_2^2\Big].
  \label{eq:flow}
\end{equation}
Because $(z_0,M)$ are geometrically aligned pairs from real drives rather than synthetic renders, minimizing $\mathcal{L}_{\text{flow}}$ learns the conditional distribution $p_{\text{data}}(X\mid M,y)$ without trading off realism against layout alignment. Crucially, all $C=7$ cameras share a single chunk noise scalar $\sigma$ per training step, preventing cross-view attention (\cref{eq:xview}) from short-circuiting geometry learning by copying features from a less-noisy neighbor.

Second, during Stage~III self-forcing distillation (\cref{eq:dmd}), both the frozen real-score teacher $v_{\text{teacher}}$ and the tracking fake-score network $v_{\text{fake}}$ evaluate generated rollouts under the identical ground-truth layout sequence $M$. The score difference therefore pushes few-step student rollouts toward the conditional real manifold $p_{\text{data}}(X\mid M,y)$ rather than an unconditional video prior.

Third, during Stage~IV Antiphon Forcing (\cref{sec:stage4}), optimizing long-horizon realism on self-generated rollouts risks subtle layout relaxation if left unconstrained. We prevent this through our \emph{Layout-Aware Kernel Score} (LAK) and dual-role architecture. Specifically, the Stage~III backbone and layout embeddings $\mathcal{F}_{\text{map}}$ remain frozen while only rank-32 role adapters are updated (\cref{eq:dual}), and conditional multi-encoder kernel matching (\cref{eq:kernel}) compares multiple generated seeds against the paired real recording sharing the same layout $M$. In addition, a frozen \emph{calibrated layout-control probe} $\mathcal{P}_\psi$ operates on dense DINOv3~\citep{oquab2025dinov3} spatial tokens over the dilated 5-class HDMap/box support (lanes, cars, trucks, pedestrians, cyclists), penalizing any excess Binary Cross-Entropy (BCE) error above the calibrated real-video VAE floor $\tau_{v,c}$:
\begin{equation}
  \mathcal{L}_{\text{ctrl}}=\frac{1}{|\mathcal{I}|}\sum_{(v,c)\in\mathcal{I}}\max\!\Big(0,\;\mathcal{E}_{\text{BCE}}\big(\mathcal{P}_\psi(\Phi_{\text{dense}}(\hat{X}_v)),\,Y^{\text{map}}_{v,c}\big)-\mathcal{E}_{\text{BCE}}\big(\mathcal{P}_\psi(\Phi_{\text{dense}}(X_v)),\,Y^{\text{map}}_{v,c}\big)-\tau_{v,c}\Big),
  \label{eq:probe}
\end{equation}
where $\Phi_{\text{dense}}$ extracts spatial DINOv3 patch tokens, $Y^{\text{map}}_{v,c}$ denotes the rasterized 5-class layout target over active view--class pairs $\mathcal{I}$, and $\tau_{v,c}\ge 0$ is calibrated on held-out VAE reconstructions to avoid penalizing irreducible codec loss. In \cref{sec:experiments}, we verify both visual fidelity and geometric controllability through monocular-SLAM ego-trajectory accuracy ($\text{ADE}/\text{DTW}/\text{FTD}$), tracked agent consistency and persistence ($\text{A-cons.}/\text{A-pres.}$), 7-edge cross-view LoFTR correspondence, and held-out layout-support similarity (\cref{fig:telemetry}).

\needspace{5\baselineskip}
\section{Training Curriculum and Antiphon Forcing}\label{sec:curriculum}

\begin{figure}[tbp]
  \centering
  \includegraphics[width=\linewidth,height=0.45\textheight,keepaspectratio]{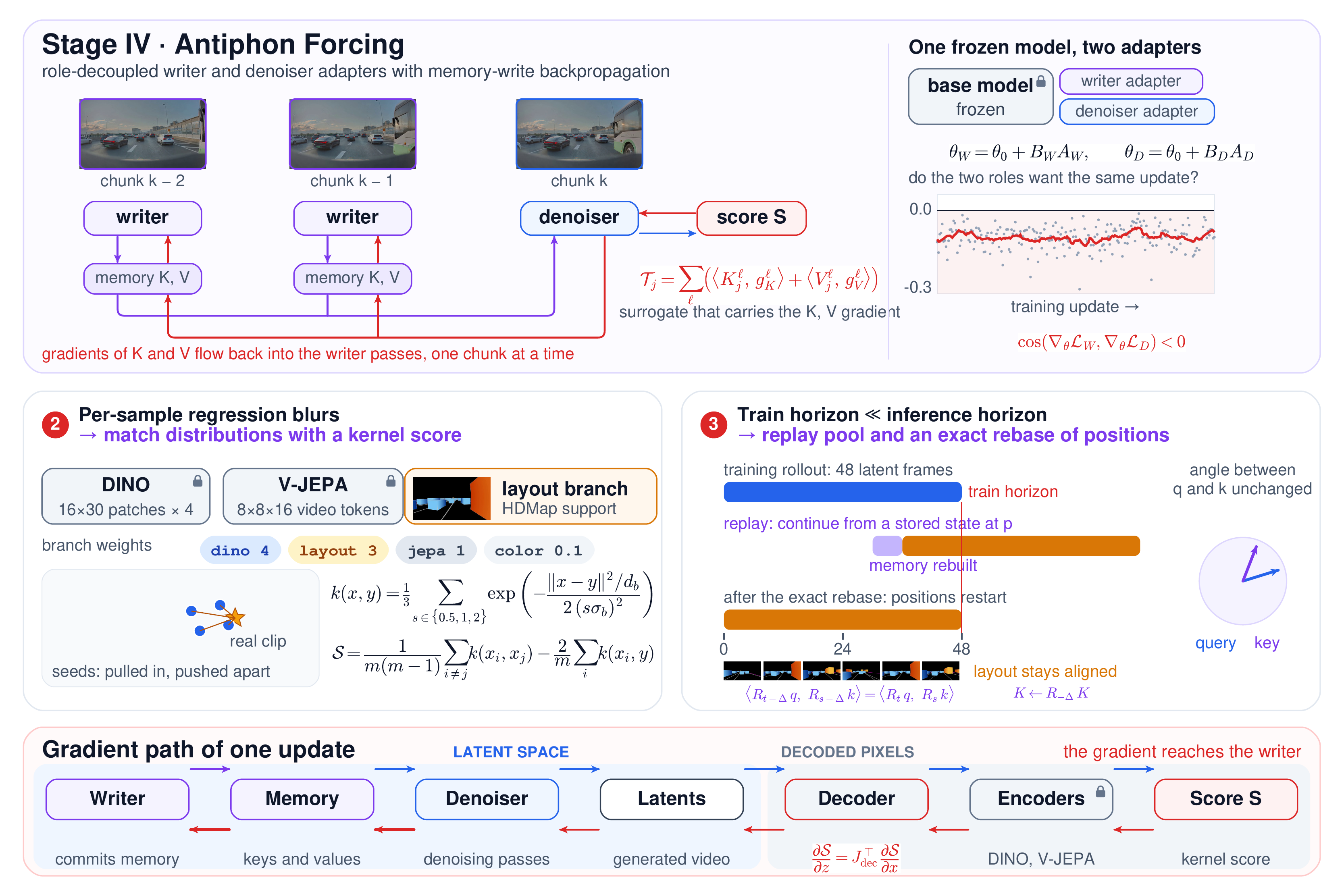}
  \caption{\textbf{Antiphon Forcing (Stage IV) architecture.} Freezing the base DiT, separate rank-32 LoRA adapters specialize the writer ($\Delta W_{\mathcal{W}}$) and denoiser ($\Delta W_{\mathcal{D}}$) roles (\cref{eq:dual}). Frozen DINOv3 and V-JEPA~2 encoders together with the calibrated HDMap layout probe supervise long-horizon rollouts (\cref{eq:kernel}), routing pixel-space gradients through the causal VAE decoder VJP into cached keys and values (\cref{eq:vjp}).}
  \label{fig:antiphon}
\end{figure}
\begin{table}[tbp]
  \centering
  \caption{\textbf{Four-stage training curriculum of BoschDreams\textsuperscript{7V}.} Training rollout length and KV-cache window are listed separately for clarity.}
  \label{tab:curriculum}
  \footnotesize
  \setlength{\tabcolsep}{3.0pt}
  \renewcommand{\arraystretch}{1.18}
  \begin{tabularx}{\linewidth}{@{}>{\RaggedRight\bfseries}p{2.35cm}>{\RaggedRight}p{2.25cm}>{\RaggedRight}X>{\RaggedRight}p{1.85cm}>{\RaggedRight}p{1.45cm}>{\RaggedRight}p{1.55cm}@{}}
    \toprule
    \hd{Stage} & \hd{Init.\ $\rightarrow$ Train} & \hd{Objective \& Mechanism} & \hd{Train Rollout} & \hd{KV Window} & \hd{Inference} \\
    \midrule
    I. Surround Teacher & 1V DiT $\rightarrow$ 7V DiT & Flow matching (\cref{eq:flow}); shared chunk $\sigma$ across 7 views & 24 latents ($9.3$\,s), bidir. & Full clip & 35 steps, CFG 5 \\
    \arrayrulecolor{boschgray!25}\midrule
    II. Causal Student & Stage~I $\rightarrow$ 7V DiT & Flow matching (\cref{eq:flow}) with 2-latent causal mask (\cref{eq:mask}) & 24 latents ($9.3$\,s), $L=2$ & 6 latents ($2.4$\,s) & 35 steps, CFG 5 \\
    \midrule
    III. Self-Forcing & Stage~II $\rightarrow$ $G_\phi$, $v_{\text{fake}}$ & Rollout DMD (\cref{eq:dmd}) on self-generated KV history & 24 latents ($9.3$\,s), $L=2$ & 6 latents ($2.4$\,s) & 1--4 steps, CFG 1 \\
    \midrule
    IV. Antiphon Forcing & Stage~III $\rightarrow$ $\Delta W_{\mathcal{W}},\Delta W_{\mathcal{D}}$ & LAK score (\cref{eq:kernel,eq:probe}) + dual LoRA (\cref{eq:dual}) + VJP (\cref{eq:vjp}) & 48 latents ($18.9$\,s) + replay & 6 latents ($2.4$\,s) & 1--4 steps, CFG 1 \\
    \arrayrulecolor{black}\bottomrule
  \end{tabularx}
\end{table}

Training a causal, few-step, seven-camera diffusion transformer directly from scratch or from single-view weights fails to converge simultaneously on cross-camera 3D geometry and autoregressive stability. BoschDreams\textsuperscript{7V} therefore decomposes learning into a four-stage curriculum (\cref{tab:curriculum,fig:curriculum,fig:antiphon}), where every stage inherits the seven-view weights of its predecessor and isolates a single modeling capability: surround geometry (Stage~I), block-causal streaming (Stage~II), few-step self-forcing distillation (Stage~III), and memory-decoupled long-horizon post-training (Stage~IV).

\subsection{Stages I--III: Surround Teacher, Causal Student, and Few-Step Distillation}\label{sec:stages123}
Stage~I initializes the 28-block backbone from a single-view HDMap diffusion transformer and trains all parameters with the Rectified Flow objective (\cref{eq:flow}) over 24-latent-frame (93-frame, $9.3$\,s) clips. As illustrated in \cref{fig:curriculum}a, drawing independent noise levels per camera allows noisy views to copy clean features from adjacent cameras via cross-view attention without learning 3D projective correspondence; sharing a single draw $\sigma\sim\mathcal{U}(0,1)$ per two-frame chunk across all seven views synchronizes feature signal-to-noise ratios across the rig.

Because bidirectional attention accesses future frames unavailable during live driving, Stage~II initializes the student from the converged Stage~I seven-view teacher and replaces full temporal attention with a two-latent-frame block-causal mask (\cref{fig:curriculum}a, bottom). To preserve the causal VAE boundary from \cref{sec:attention} (where latent frame $0$ encodes exclusively initial RGB frame $0$), latent $0$ forms its own conditioning block ($b(0)=0$) and subsequent generated latents $t\ge 1$ are grouped in pairs $b(t)=\lceil t/2\rceil$:
\begin{equation}
  \mathbf{M}_{\text{causal}}(q,k)=\begin{cases}0,&b(k)\le b(q),\\-\infty,&\text{otherwise},\end{cases}\qquad b(t)=\begin{cases}0,&t=0,\\\lceil t/2\rceil,&t\ge 1.\end{cases}
  \label{eq:mask}
\end{equation}
Retaining full-backbone trainability under \cref{eq:flow} adapts temporal representations to causal execution while preserving the learned 14-directed-edge cross-camera geometry.

Stage~III compresses sampling from 35 guided steps (70 NFE) to $N_{\text{step}}\in\{1,2,4\}$ unguided passes (\cref{fig:curriculum}b). Rather than distilling on ground-truth prefixes, the few-step generator $G_\phi$ unrolls future chunks $\hat{z}^{(k)}$ autoregressively on its own generated KV cache, committing each finished chunk via a light-noise write pass ($\sigma_{\text{write}}=0.13$). Re-noising $\hat{z}^{(k)}$ to $z_\sigma$, regression-free Distribution Matching Distillation (DMD/DMD2)~\citep{yin2024dmd,yin2024dmd2} under Self Forcing~\citep{huang2025selfforcing} updates $G_\phi$ using the score difference between the frozen Stage~I CFG teacher $v_{\text{teacher}}$ and a tracking fake-score network $v_{\text{fake}}$:
\begin{equation}
  \nabla_\phi\mathcal{L}_{\text{DMD}}=\E_{\sigma,\epsilon,M,y}\Big[\omega(\sigma)\big(\sg[v_{\text{teacher}}(z_\sigma,\sigma,M,y)]-\sg[v_{\text{fake}}(z_\sigma,\sigma,M,y)]\big)\frac{\partial G_\phi}{\partial\phi}\Big],
  \label{eq:dmd}
\end{equation}
where $\sg[\cdot]$ denotes stop-gradient and $\omega(\sigma)$ weights flow noise levels. Classifier-free guidance is absorbed into $G_\phi$, enabling single-branch inference ($\text{CFG}=1$).

\subsection{Stage IV: Antiphon Forcing and Dual-Role Memory Decoupling}\label{sec:stage4}
\begin{figure}[tbp]
  \centering
  \includegraphics[width=\linewidth,height=0.22\textheight,keepaspectratio]{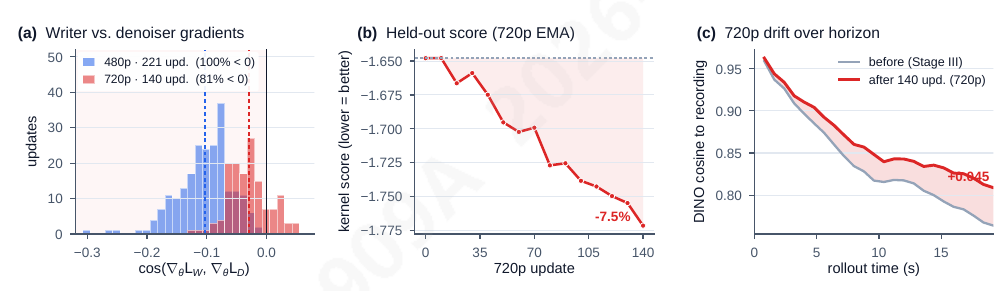}
  \caption{\textbf{Stage IV training telemetry.} (a) Per-step cosine similarity between writer and denoiser gradients across 480p (221 steps, $100\%$ negative) and 720p (140 steps, $81\%$ negative) runs. (b) Held-out 720p EMA feature-kernel score ($16$ conditions $\times 2$ seeds; lower is better). (c) 720p DINOv3 cosine similarity to recorded video over the 48-latent ($18.9$\,s) rollout horizon before and after Stage~IV.}
  \label{fig:telemetry}
\end{figure}

Although Stage~III achieves real-time few-step generation, rollouts beyond 10--20 chunks accumulate texture softening and exposure drift. Consistent with the role-antagonism phenomenon identified in single-view autoregressive video generation by SGF+~\citep{su2026sgfplus}, per-role gradient analysis during seven-camera streaming reveals a clear optimization conflict (\cref{fig:antiphon,fig:telemetry}a). Every chunk executes two distinct roles: a \emph{denoiser role} ($\mathcal{D}$), which maps shifted flow noise levels $\{1.00, 0.94, 0.83, 0.63\}$~\citep{esser2024sd3} to clean velocities, and a \emph{memory-writer role} ($\mathcal{W}$), which encodes completed latents at $\sigma_{\text{write}}=0.13$ into persistent KV cache states $K^{(k)},V^{(k)}$ for future chunks. When both roles share one parameter vector $\theta$, their gradients systematically oppose each other: $\cos(\nabla_\theta\mathcal{L}_{\mathcal{W}},\nabla_\theta\mathcal{L}_{\mathcal{D}})$ lies in $[-0.302,-0.012]$ with mean $-0.103$ across all 221 steps at 480p ($100\%$ negative), and is negative on $81\%$ of the 140 steps at 720p (mean $-0.030$). Updating shared weights therefore trades immediate denoising sharpness against long-horizon memory fidelity.

\emph{Antiphon Forcing} resolves this conflict in our seven-camera layout-conditioned setting through three coupled mechanisms (\cref{fig:antiphon}):
\begin{enumerate}
  \item \textbf{Dual-role LoRA decoupling:} Adapting role-specific writer/denoiser parameterization~\citep{su2026sgfplus} without duplicating the full seven-camera backbone, we freeze the Stage~III weights $W_{\text{base}}$ and equip each attention projection with separate rank-32 low-rank adapters~\citep{hu2021lora} for writing and denoising:
  \begin{equation}
    W=W_{\text{base}}+\Delta W_{\mathcal{W}}\,\mathbf{1}_{[\text{role}=\mathcal{W}]}+\Delta W_{\mathcal{D}}\,\mathbf{1}_{[\text{role}=\mathcal{D}]},\qquad \Delta W_{\mathcal{W}}=B_{\mathcal{W}}A_{\mathcal{W}},\ \Delta W_{\mathcal{D}}=B_{\mathcal{D}}A_{\mathcal{D}}.
    \label{eq:dual}
  \end{equation}
  Activating $\Delta W_{\mathcal{W}}$ exclusively during KV-cache commits and $\Delta W_{\mathcal{D}}$ during denoising steps eliminates inter-role gradient cancellation on shared weights.
  \item \textbf{Layout-Aware Kernel Score (LAK):} Stage~IV unrolls 48-latent-frame ($18.9$\,s, 189-frame) surround rollouts, while a FIFO continuation replay pool extends supervision beyond the training horizon. Frozen perceptual encoders $b\in\{\text{DINOv3},\text{V-JEPA~2},\text{color}\}$~\citep{oquab2025dinov3,assran2025vjepa2} embed decoded video windows alongside dense DINOv3 tokens on the dilated HDMap support. Using a multi-scale Gaussian kernel $k_b(x,y)=\frac{1}{3}\sum_{s\in\{0.5,1,2\}}\exp(-\|x-y\|_2^2/(2\,d_b\,s^2\sigma_b^2))$ normalized by feature dimension $d_b$ with calibrated bandwidth $\sigma_b$~\citep{gneiting2007scoring,gretton2012mmd}, LAK attracts generated seeds $g$ toward the paired recording $r$ while repelling distinct seeds $g\neq g'$ under the identical layout $M$:
  \begin{equation}
    \mathcal{L}_{\text{LAK}}=\sum_{b}w_b\Big(\E_{g\neq g'}\big[k_b(g,g')\big]-2\,\E_{g}\big[k_b(g,r)\big]\Big)+\lambda_{\text{ctrl}}\,\mathcal{L}_{\text{ctrl}}+\lambda_{\text{anc}}\,\|\hat{z}_{1:2}-z^{\text{base}}_{1:2}\|_2^2.
    \label{eq:kernel}
  \end{equation}
  We set branch weights $(w_{\text{DINOv3}},w_{\text{V-JEPA~2}},w_{\text{color}})=(4.0,1.0,0.1)$ and layout-probe weight $\lambda_{\text{ctrl}}=3.0$ (\cref{fig:antiphon}). Here $\mathcal{L}_{\text{ctrl}}$ is the calibrated 5-class HDMap/agent probe (\cref{eq:probe}), and the anchor term ($\lambda_{\text{anc}}=0.30$ at 480p; $0.05$ at 720p) pins the opening two chunks to the frozen Stage~III parent $z^{\text{base}}$ to preserve short-horizon sharpness. Training uses AdamW (peak LR $1\times 10^{-5}$, gradient clip $1.0$, population size $128$ across 32 GPUs, with $m=4$ seeds/condition at 480p and $m=2$ at 720p; \cref{tab:config}). LAK alone (shared adapter in \cref{tab:longhorizon,tab:ablation}) provides Stage~IV perceptual and structural supervision; combining LAK with dual-role adapters and memory VJP yields full \emph{Antiphon Forcing}.
  \item \textbf{Decoder-VJP temporal memory backpropagation:} Rather than detaching cached keys and values, downstream cotangents from chunk $k{+}1$ are mapped from pixel space back to latent space via the exact Vector--Jacobian Product (VJP) of the causal VAE decoder ($g_{\text{latent}}=\mathbf{J}_{\mathcal{D}_{\text{VAE}}}^{\top}g_{\text{pixel}}$) and backpropagated through the replayed writer passes of preceding chunks ($H=2$):
  \begin{equation}
    \frac{\partial\mathcal{L}_{\text{LAK}}^{(k+1)}}{\partial\Delta W_{\mathcal{W}}^{(k)}}=\frac{\partial\mathcal{L}_{\text{LAK}}^{(k+1)}}{\partial\hat{z}^{(k+1)}}\cdot\frac{\partial\hat{z}^{(k+1)}}{\partial(K^{(k)},V^{(k)})}\cdot\frac{\partial(K^{(k)},V^{(k)})}{\partial\Delta W_{\mathcal{W}}^{(k)}}.
    \label{eq:vjp}
  \end{equation}
\end{enumerate}

\begin{figure}[tbp]
  \centering
  \includegraphics[width=\linewidth,height=0.30\textheight,keepaspectratio]{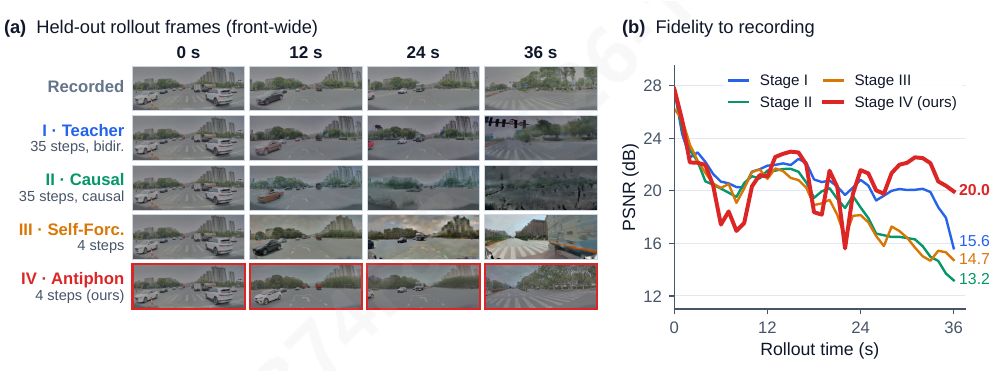}
  \caption{\textbf{Qualitative 36\,s held-out rollout across curriculum Stages I--IV.} (a) Front-wide frames at $0,12,24,36$\,s under identical initial frame, 7-camera HDMap/box stream, and caption (Stage~I extended via sliding-window re-conditioning past its $9.3$\,s bidirectional window). (b) Per-second PSNR against the recorded drive on this single-scene case study.}
  \label{fig:stage_progression}
\end{figure}

As shown in \cref{fig:telemetry}b--c, 140 training steps of 720p Antiphon Forcing improve the held-out EMA feature-kernel score by $7.5\%$ and raise $18.9$\,s ($48$-latent) DINOv3 similarity to the recording from $0.764$ to $0.809$ ($+0.045$) without altering initial-frame fidelity. On a qualitative $36$\,s held-out rollout case study (\cref{fig:stage_progression}, where Stage~I is extended via sliding-window re-conditioning across $9.3$\,s segments because its bidirectional attention cannot extrapolate natively), Stage~I, Stage~II, and Stage~III degrade to $13.2$--$15.6$\,dB PSNR at $36$\,s, whereas Stage~IV sustains crisp road and agent structure at $20.0$\,dB PSNR (population-level 600-scene quantitative ablations are reported in \cref{sec:ablation_results}).

\needspace{5\baselineskip}
\section{Real-Time Streaming Runtime and Closed-Loop AI Agent}\label{sec:runtime}

\subsection{Chunked Rollout Execution and Exact Rotary Position Rebase}\label{sec:rebase}
\begin{figure}[!tbp]
  \centering
  \includegraphics[width=\linewidth,height=0.50\textheight,keepaspectratio]{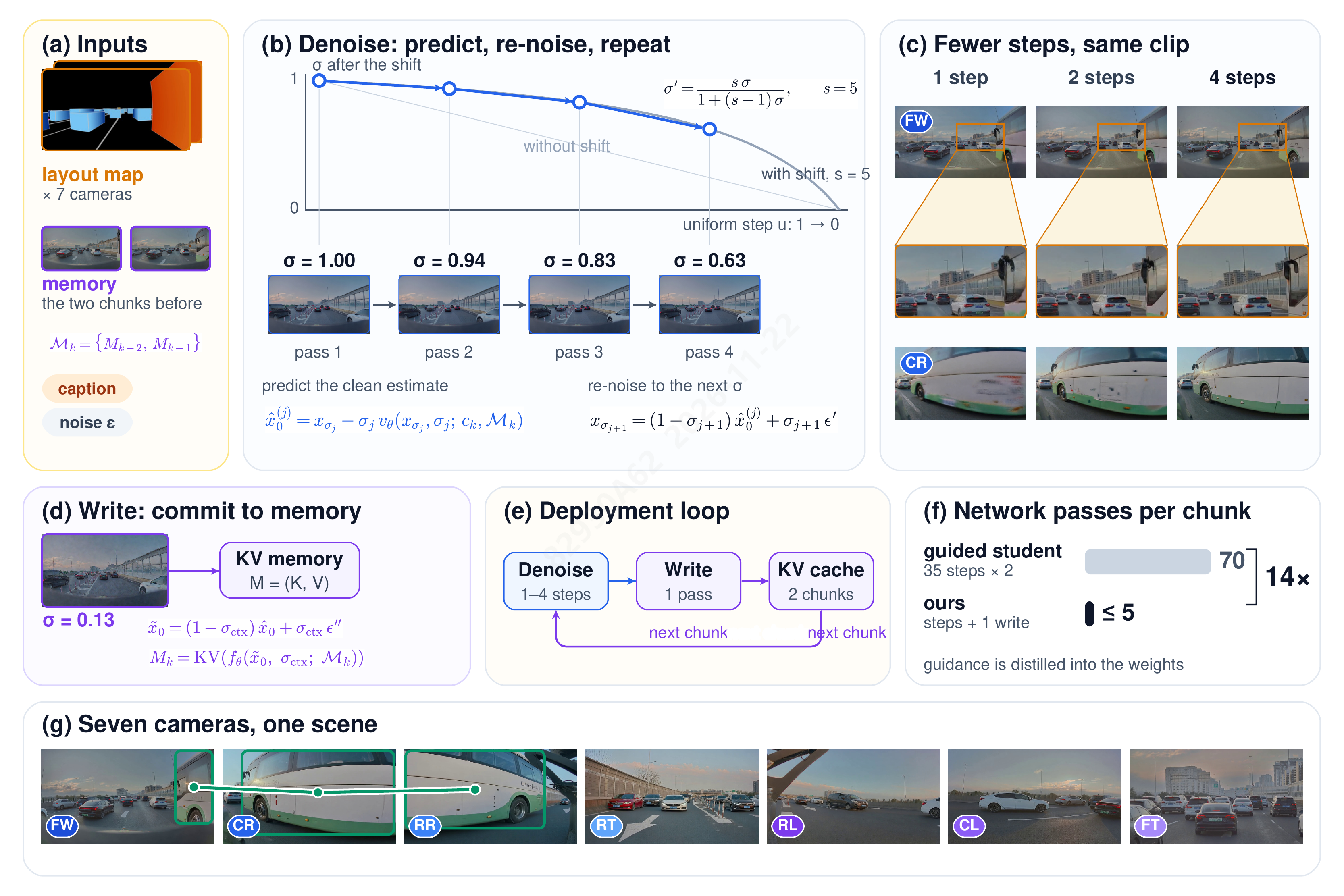}
  \vspace{-4mm}
  \caption{\textbf{Chunk-wise streaming execution and KV-memory lifecycle.} (a) Each 7-camera chunk ($8$ frames $\times 7$ views) starts from Gaussian noise conditioned on projected layouts, bounded 6-latent KV cache, and caption. (b--c) Shifted flow-matching steps synthesize clean latents in 1, 2, or 4 passes. (d) The writer role commits the finished chunk to KV memory at $\sigma_{\text{write}}=0.13$. (e--f) Distilled $\text{CFG}=1$ removes the dual-branch teacher KV cache ($70\to 1$--$4$ denoising passes), outputting synchronized 7-view streams (g).}
  \label{fig:rollout}
\end{figure}

In streaming deployment, BoschDreams\textsuperscript{7V} synthesizes video in causal chunks of $L=2$ latent frames ($8$ display frames per camera at $10$\,Hz, spanning an $800$\,ms simulation tick). As depicted in \cref{fig:rollout}, each chunk follows a deterministic static-shape execution model~\citep{nvidia2026omnidreams}. First, Gaussian noise is initialized alongside projected 7-camera HDMap/box layouts, cached KV memory, and the scene prompt (a), and denoised in $N_{\text{step}}\in\{1,2,4\}$ shifted flow-matching passes (b--c). Next, a single memory-write pass ($\Delta W_{\mathcal{W}}$ at $\sigma_{\text{write}}=0.13$) commits the finished chunk into a sliding six-latent-frame KV window—the active chunk plus two preceding chunks—bounding memory strictly in time (d). Because distilled single-branch execution ($\text{CFG}=1$) removes the dual-branch teacher KV cache (e--f), asynchronous LightTAE~\citep{lightx2v2025,madebyollin2025taehv} decodes synchronized seven-view streams while the DiT immediately begins the next chunk (g).

\begin{figure}[ht]
  \centering
  \includegraphics[width=\linewidth,height=0.28\textheight,keepaspectratio]{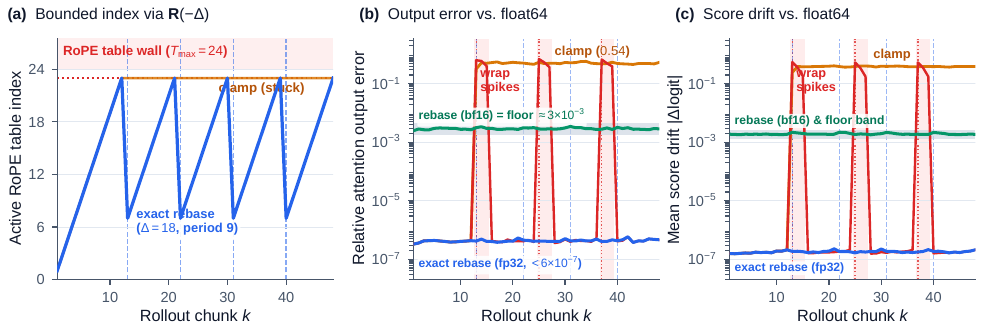}
  \vspace{-5mm}
  \caption{\textbf{RoPE rebase probe ($T_{\max}=24$, 6-latent KV window).}
  (a) Active RoPE index under clamping, modulo wrapping, and pre-emptive $\mathbf{R}(-\Delta)$ rebasing ($\Delta{=}18$, period 9).
  (b, c) Output error and logit drift vs.\ a float64 reference.
  Blue dashed lines: rebase triggers (chunks 13, 22, 31, 40); red bands: modulo wrap across the table boundary (chunks 13, 25, 37).
  Rebasing stays at the bf16 floor ($\approx 3\times10^{-3}$); fp32 rebasing stays below $6\times10^{-7}$.}
  \label{fig:rebase}
\end{figure}

\begin{figure}[tbp]
  \centering
  \includegraphics[width=\linewidth,height=0.48\textheight,keepaspectratio]{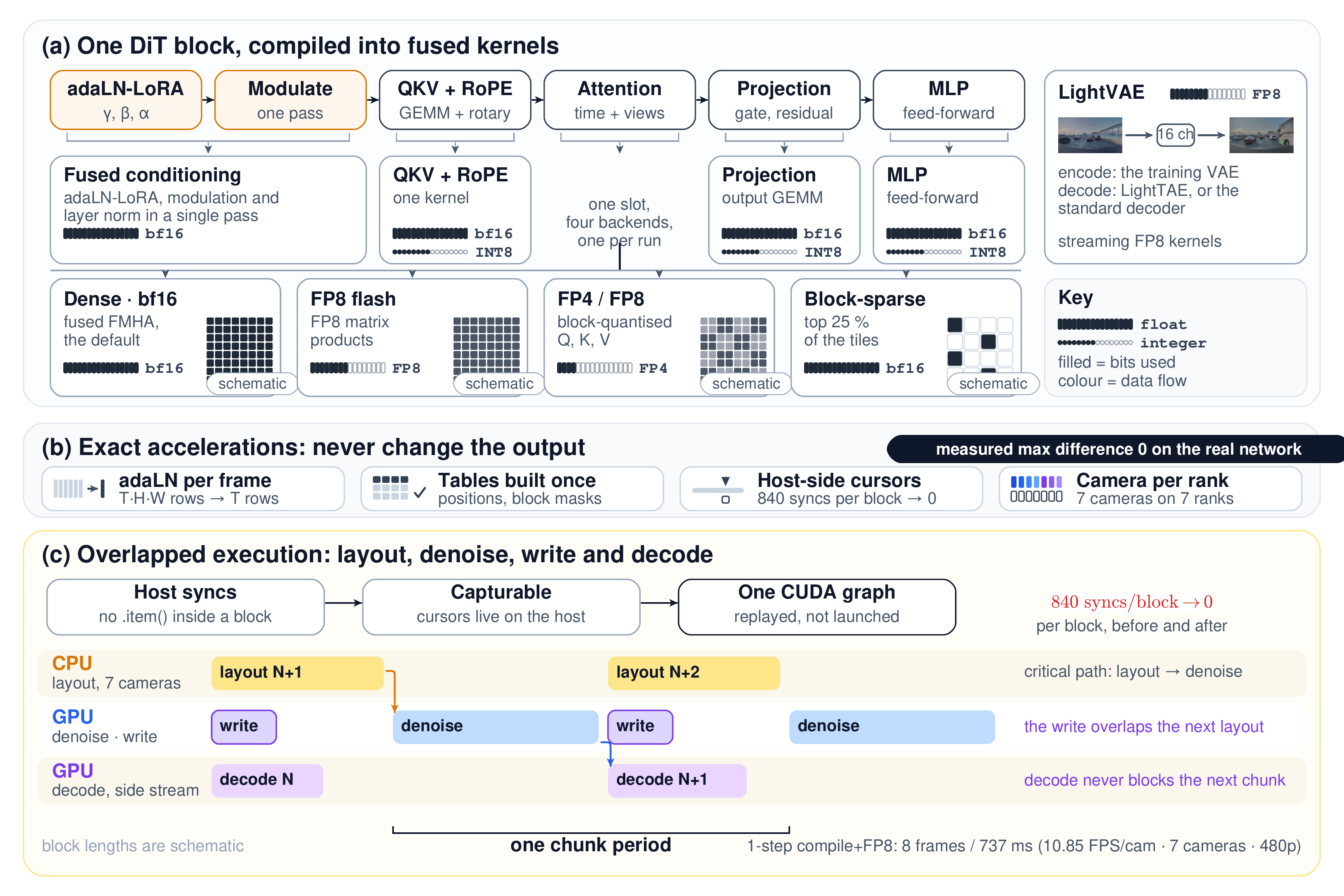}
  \caption{\textbf{Fused streaming block architecture and asynchronous CUDA Graph pipeline.} (a) Single-pass frame-granular \texttt{adaLN}, fused QKV projection with in-kernel RoPE, and pluggable FlashAttention/FP8 backends. (b) Numerically exact speedups verified at exactly zero relative error against the eager reference. (c) Full-model CUDA Graph capture overlapped with asynchronous CPU layout rasterization and LightTAE decoding (1-step compile+FP8: $737$\,ms / 8 frames, $10.85$\,FPS per camera at 480p).}
  \label{fig:engineering}
\end{figure}

Because 1D temporal RoPE frequencies are pre-tabulated for a finite horizon $T_{\max}=128$ latent frames ($51.2$\,s), naive autoregressive indexing fails once rollout length exceeds $T_{\max}$. Let $s$ denote the next global latent start index, $o$ the current logical origin offset, and $W=6$ the retained KV window length. Exploiting the shift invariance $\mathbf{R}(q)^{\top}\mathbf{R}(k)=\mathbf{R}(k-q)$, whenever the next chunk would cross $T_{\max}$, the runtime pre-emptively rotates all cached keys in place by $-\Delta=-(s-o-W)$ in fp32 (storing back in bf16) and shifts query and key frame indices downward by $\Delta$:
\begin{equation}
  \tilde{K}_{\text{cached}}=K_{\text{cached}}\,\mathbf{R}(-\Delta),\qquad \big\langle\mathbf{R}(q-\Delta)Q,\;\mathbf{R}(k-\Delta)K\big\rangle=\big\langle\mathbf{R}(q)Q,\;\mathbf{R}(k)K\big\rangle.
  \label{eq:rebase}
\end{equation}
\Cref{fig:rebase} benchmarks this exact rotary rebase against an unbounded float64 reference across 48 chunks under a $T_{\max}=24$ stress test ($W=6$ latent-frame KV window): because rebasing triggers \emph{before} table exhaustion to keep the entire 6-frame window inside $[0,T_{\max}-1]$, its origin shifts recur every $(T_{\max}-W)/2=9$ chunks (chunks $13,22,31,40$), whereas naive modulo wrapping straddles the $23\to 0$ table end every $T_{\max}/2=12$ chunks (red shaded intervals at chunks $13,25,37$) and index clamping saturates permanently ($\approx 0.54$ error). Exact rebase maintains $<6\times 10^{-7}$ error in fp32 and sits on the $\approx 3\times 10^{-3}$ bf16 storage floor across extended horizons.

\subsection{Zero-Synchronization Kernel Fusion and Real-Time Efficiency}\label{sec:efficiency}
To reach the $10$\,Hz camera clock, the runtime pairs three numerically exact ($0.00$ relative error vs.\ eager reference) structural optimizations with compiled low-precision attention backends (\cref{fig:engineering}): (i) \emph{frame-granular \texttt{adaLN}}, evaluating the $(\sigma,v)$ modulation MLP once per camera frame and broadcasting across all $N_s$ spatial tokens (removing $99.9\%$ of modulation FLOPs at $0.00$ relative error); (ii) \emph{precomputed RoPE lookup tables} indexed by integer offsets ($0.00$ relative error); and (iii) \emph{host-side 64-bit KV cache cursors}, removing $\sim 840$ device-to-host scalar synchronization stalls per block so the entire 28-block seven-camera forward pass captures into a single CUDA Graph alongside fused QKV/RoPE kernels, \texttt{torch.compile}, and pluggable exact or quantized/sparse attention backends (FlashAttention-2/3~\citep{dao2023flashattention2,shah2024flashattention3}, SageAttention3~\citep{zhang2025sageattention3}, and SpargeAttention~\citep{zhang2025spargeattn}).

\begin{figure}[tbp]
  \centering
  \includegraphics[width=\linewidth,height=0.32\textheight,keepaspectratio]{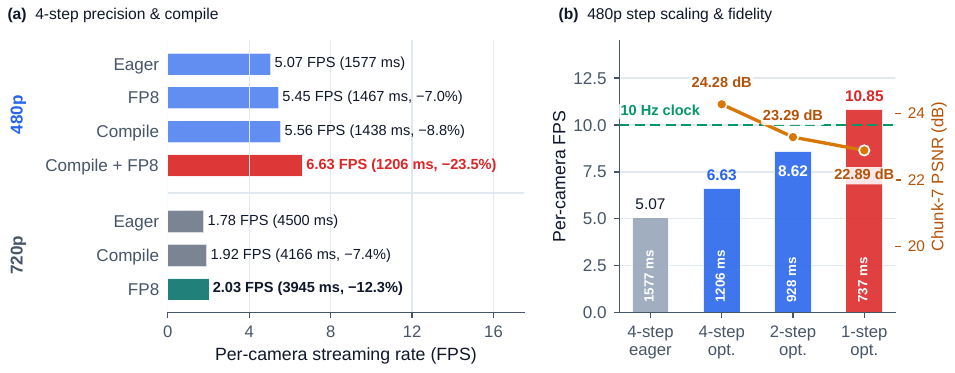}
  \caption{\textbf{Measured streaming efficiency and step-scaling trade-off.} (a) 4-step precision and compilation ablation at 480p and 720p (wall-clock ms per 8-frame 7-camera chunk and per-camera FPS). (b) Sampling-step scaling at 480p (\texttt{compile+FP8}) vs.\ the $10$\,Hz real-time camera clock and chunk-7 PSNR on the held-out reference drive.}
  \label{fig:latency}
\end{figure}

We report end-to-end steady-state streaming rate strictly as \emph{per-camera FPS}, defined as $\text{FPS}=8/(\text{wall}_{\text{ms}}/1000)$, the synchronous frame rate at which all seven camera streams advance together (full numeric breakdown and auxiliary 7-camera aggregate throughput $\text{FPS}_{\text{agg}}=7\times\text{FPS}$ are tabulated in \cref{sec:system,tab:supp_efficiency}). As shown in \cref{fig:latency}a, at 4 denoising steps, \texttt{torch.compile} and FP8 attention compose to cut 480p per-chunk wall-clock latency by $23.5\%$ ($1577\to 1206$\,ms, reaching $6.63$\,FPS, $1.31\times$ over eager); at 720p, FP8 attention alone reduces latency by $12.3\%$ ($4500\to 3945$\,ms, $2.03$\,FPS, $1.14\times$ over eager), while \texttt{torch.compile} alone reaches $4166$\,ms ($1.92$\,FPS). Scaling sampling steps under the compiled FP8 runtime at 480p (\cref{fig:latency}b) reduces wall-clock chunk time to $928$\,ms at 2 steps ($8.62$\,FPS, $-41.2\%$ vs.\ 4-step eager) and $737$\,ms at 1 step, achieving \textbf{10.85\,FPS} ($\approx 11$\,FPS)—a $2.14\times$ speedup over the 4-step eager baseline ($5.07$\,FPS) that surpasses the $10$\,Hz physical sensor clock at a $1.39$\,dB trade-off in chunk-7 PSNR ($24.28\to 22.89$\,dB across the 4/2/1-step compiled FP8 checkpoints; eager 4-step matches compiled 4-step numerically). Beyond standard GPUs, runtime compatibility shims also support deployment on domestic T-Head PPU and Hygon DCU accelerators (\cref{tab:capability}).

\subsection{Closed-Loop Simulation and Autonomous AI-Agent Orchestration}\label{sec:closedloop}
\begin{figure}[tbp]
  \centering
  \includegraphics[width=\linewidth,height=0.50\textheight,keepaspectratio]{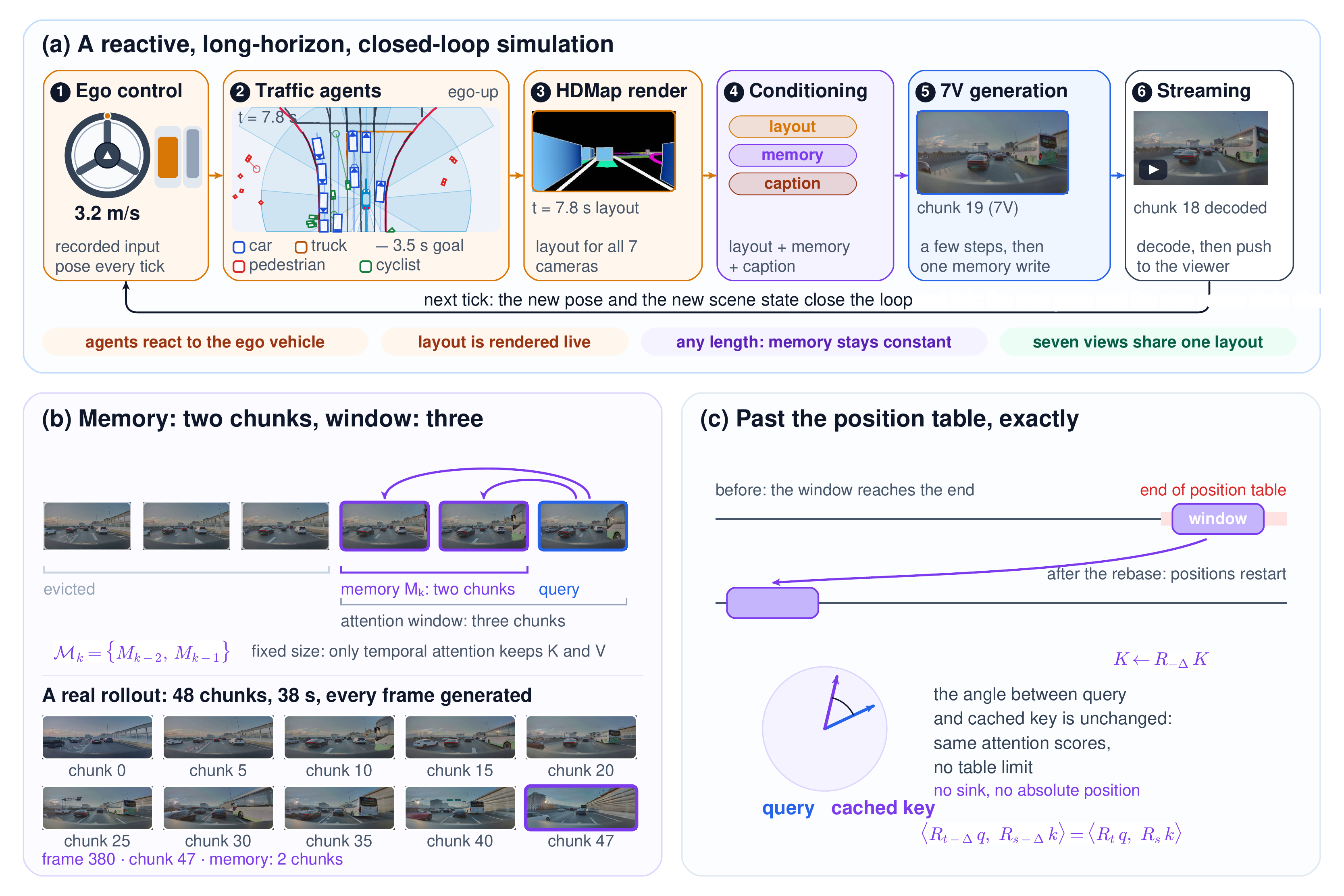}
  \caption{\textbf{Closed-loop counterfactual simulation and rolling-memory architecture.} (a) External ego planner or AI-agent controls update the 3D BEV traffic simulator, which projects reactive agent boxes and HDMap vectors into all seven cameras every $0.8$\,s chunk. (b) Sliding 6-latent-frame KV window bounds GPU memory over continuous rollouts. (c) Exact in-place RoPE key rotation (\cref{eq:rebase}) rebases cached positions past the finite table boundary.}
  \label{fig:closedloop}
\end{figure}

Because traffic state evolution is decoupled from pixel rendering (\cref{fig:closedloop}), BoschDreams\textsuperscript{7V} closes the perception--planning--simulation loop without retraining the generator. The simulator integrates a kinematic bicycle model~\citep{kong2015bicycle} and Intelligent Driver Model (IDM)~\citep{treiber2000idm} agent dynamics internally at $10$\,Hz, while policy control updates are applied at every $800$\,ms chunk boundary ($1.25$\,Hz closed-loop control rate, with first-chunk rendering latency of $\approx 737$\,ms plus queueing overhead at 480p 1-step; throughput exceeds real-time sensor rate). At each chunk boundary, updated 3D boxes and HDMap vectors are projected into all seven cameras while rolling KV memory and exact RoPE rebasing sustain multi-minute horizon continuity (\cref{fig:closedloop}b--c).

Building on this controllable interface, we deploy an end-to-end \emph{autonomous AI scenario agent} that converts real-world fleet failure tickets directly into closed-loop training and verification worlds with zero human labeling. Given a fleet disengagement or corner-case ticket, the CPU-side AI agent parses the failure semantics, retrieves matching base drives, edits the 3D vector world to synthesize counterfactual traffic branches (e.g., occluded pedestrian jaywalking, aggressive cut-ins, or multi-branch merge decisions), and verifies kinematic plausibility before BoschDreams\textsuperscript{7V} renders synchronized seven-camera streams for planner retraining and closed-loop pass/fail regression testing.

\needspace{5\baselineskip}
\section{Diffusion Fixer for 3D Gaussian Splatting}\label{sec:fixer}

Explicit 3D Gaussian Splatting (3DGS)~\citep{kerbl2023gaussians,chen2025omnire} enables real-time viewpoint manipulation of logged drives, yet novel ego trajectories expose systematic failure modes: sparse Gaussian support produces semi-transparent floaters and missing background structures, non-rigid dynamic actors smear across frames, and rasterized splats lack realistic camera optics (rolling-shutter blur, lens flare, sensor noise). Building on the paired generative restoration paradigm explored in single-step and online reconstruction enhancers~\citep{wu2025difix,zhang2026diffusionharmonizer,delutio2026artifixer} as well as the paired Diffusion Fixer in OmniDreams~\citep{nvidia2026omnidreams}, we adapt our seven-camera HDMap backbone into a \emph{multi-view Diffusion Fixer} that transports degraded 3DGS renders onto the distribution of real surround camera video while preserving 3D layout alignment.

\subsection{Well-Posedness of the Stochastic Paired Flow Bridge}\label{sec:bridge}
Let $z_{\text{src}}$ and $z_{\text{gt}}$ denote the causal VAE latents of a degraded 7-camera 3DGS rendering and its paired real recording, with 32-channel conditioning $c=[z_{\text{layout}}\,|\,z_{\text{src}}]$ concatenating the projected 3D HDMap/box layout and degraded latent. Regressing $z_{\text{gt}}$ deterministically blurs high-frequency road and foliage textures, whereas naive noise-free linear interpolation between $z_{\text{src}}$ and $z_{\text{gt}}$ suffers a fundamental mathematical degeneracy. Related to stochastic interpolants and image-to-image Schrödinger bridges~\citep{albergo2023interpolants,liu2023i2sb}, we construct a \emph{stochastic paired rectified-flow bridge} by perturbing the degraded endpoint with calibrated Gaussian noise $\sigma_{\text{src}}=0.3$:
\begin{equation}
  x_1=z_{\text{src}}+\sigma_{\text{src}}\epsilon,\quad x_0=z_{\text{gt}},\quad x_t=t\,x_1+(1-t)\,x_0,\quad \mathcal{L}_{\text{fix}}=\E_{t,\epsilon}\big\|v_\theta(x_t,t,c)-(x_1-x_0)\big\|_2^2,
  \label{eq:bridge}
\end{equation}
where $\epsilon\sim\Normal(0,\mathbf{I})$ and $t\sim\mathcal{U}(0,1)$, integrated from $t=1\to 0$ in 8 fp32 Euler steps at inference.

\begin{figure}[tbp]
  \centering
  \includegraphics[width=\linewidth,height=0.28\textheight,keepaspectratio]{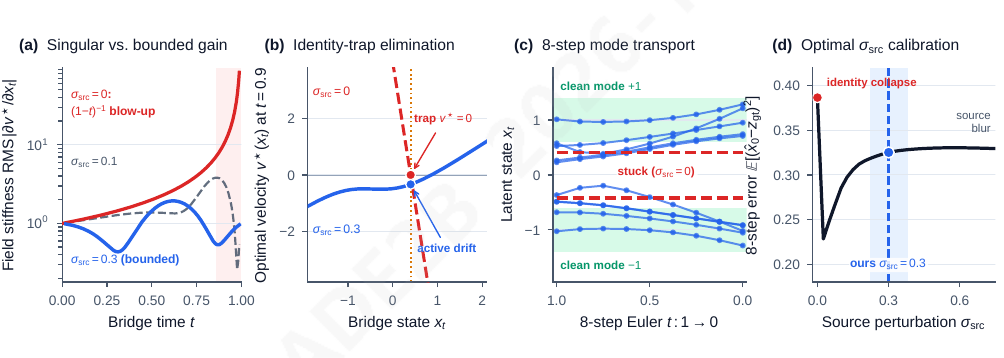}
  \caption{\textbf{Stochastic source perturbation ($\sigma_{\text{src}}=0.3$) in the paired flow-matching bridge.} Exact closed-form Bayes-optimal fields on a bimodal latent prior: (a) RMS field stiffness vs.\ $t$; (b) velocity field at $t=0.9$; (c) 8-step Euler trajectories ($t:1\to 0$); (d) expected 8-step reconstruction error $\E[(\hat{x}_0-z_{\text{gt}})^2]$ across $\sigma_{\text{src}}$.}
  \label{fig:sigma}
\end{figure}

\paragraph{Why source noise $\sigma_{\text{src}}>0$ is necessary—and why $\sigma_{\text{src}}=0.3$ is optimal.}
Because $x_1=(x_t-(1-t)z_{\text{gt}})/t$, the Bayes-optimal velocity field under \cref{eq:bridge} is $v^\star(x_t,t,c)=\big(x_t-\E[z_{\text{gt}}\mid x_t,c]\big)/t$. In a deterministic bridge ($\sigma_{\text{src}}=0$), conditioning on $z_{\text{src}}$ inside $c$ makes $z_{\text{gt}}=(x_t-t\,z_{\text{src}})/(1-t)$ algebraically recoverable from $(x_t,c)$ for every $t<1$. Consequently, the optimal field collapses to $v^\star(x_t,t,c)=(z_{\text{src}}-x_t)/(1-t)$: its spatial Jacobian $\partial v^\star/\partial x_t=-(1-t)^{-1}$ diverges near $t=1$ (\cref{fig:sigma}a), and at the inference start point $x_1=z_{\text{src}}$ the velocity vanishes ($v^\star(z_{\text{src}},t,c)\equiv 0$), creating a degenerate fixed-point trap that leaves degraded inputs untouched (\cref{fig:sigma}b--c). Injecting $\sigma_{\text{src}}>0$ breaks algebraic invertibility by introducing an irreducible Gaussian observation bandwidth $(t\sigma_{\text{src}})^2$: small $\sigma_{\text{src}}$ leaves residual stiffness near $t=1$, whereas overly large $\sigma_{\text{src}}$ erases fine source scene structure, producing a clear U-shaped 8-step transport optimum around our calibrated $\sigma_{\text{src}}=0.3$ (\cref{fig:sigma}d).

\subsection{Architecture, Restoration Fidelity, and BEV Manifold Alignment}\label{sec:fixer_results}
Initialized from the Stage~I seven-camera teacher, the Fixer ingests the additional 16 degraded-source channels via a zero-initialized patch-embedding extension—preserving exact teacher behavior at step $0$ while enabling immediate gradient flow—and trains across $6{,}003$ paired seven-camera clips spanning five canonical 3DGS degradation families: \emph{chunk seams}, \emph{under-fitting}, \emph{missing background}, \emph{foreground splat noise}, and \emph{road-surface noise}. Conditioning-frame masks are zeroed throughout so no ground-truth RGB frame can leak into restoration. Long drives are restored in 29-frame windows tiled with 4-frame linear latent cross-fading.

\begin{figure}[tbp]
  \centering
  \includegraphics[width=\linewidth,height=0.38\textheight,keepaspectratio]{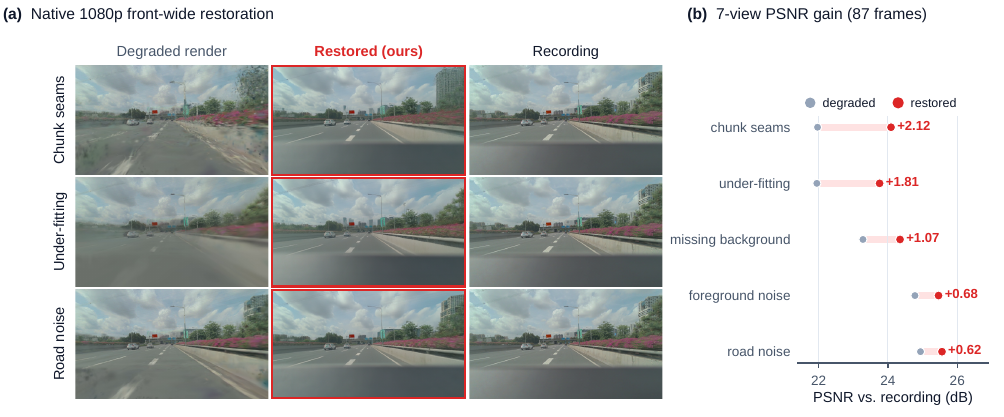}
  \caption{\textbf{Multi-view Diffusion Fixer on 3DGS degradations.} (a) Native $1920\times 1080$ \texttt{FW} restorations (8 Euler steps, $\sigma_{\text{src}}=0.3$) alongside degraded 3DGS inputs and real recordings. (b) Seven-camera, 87-frame PSNR improvements across all five 3DGS degradation families.}
  \label{fig:restoration}
\end{figure}

\Cref{fig:restoration,tab:fixer} evaluate 8-step surround restoration across all seven cameras and 87 frames on a dedicated 3DGS evaluation split at native vehicle camera resolution ($1920\times 1080$ for \texttt{FW}/\texttt{FT}; $1280\times 720$ for side/rear views). We report pixel reconstruction ($\text{PSNR}\uparrow$), structural similarity ($\text{SSIM}\uparrow$)~\citep{wang2004ssim}, perceptual distance ($\text{LPIPS}\downarrow$, AlexNet)~\citep{zhang2018lpips}, image and video distribution fidelity ($\text{FID}\downarrow$~\citep{heusel2017fid}, $\text{FVD}\downarrow$~\citep{unterthiner2018fvd}), and multi-camera 3D representation alignment via \textbf{Fr\'echet BEV Distance} ($\text{FBD}\downarrow$), computed in the feature space of a frozen \textbf{BEVFormer}~\citep{li2022bevformer} encoder.

Three clear patterns emerge in \cref{tab:fixer}. First, the largest pixel, perceptual, and video-distribution gains occur on severe structural 3DGS failures. On chunk seams, the Fixer delivers \textbf{$+2.12$\,dB PSNR}, $+0.0258$ SSIM, \textbf{$-29.5\%$ LPIPS} ($0.4201\to 0.2961$), \textbf{$-38.0\%$ FID} ($5.13\to 3.18$), and \textbf{$-45.7\%$ FVD} ($280.3\to 152.1$); on under-fitted geometry, it achieves \textbf{$+1.81$\,dB PSNR}, \textbf{$-36.1\%$ FID} ($4.76\to 3.04$), and \textbf{$-35.3\%$ FVD} ($261.4\to 169.2$), as 14-directed-edge cross-view attention and causal temporal priors reconstruct coherent boundaries where splat density collapses. Second, mild splat degradations exhibit a classic perception--distortion trade-off on SSIM: where smooth Gaussian alpha-blending yields high local luminance correlation, transporting latents onto the sharper real-sensor manifold trades $<0.008$ SSIM for $10\%$--$14\%$ lower LPIPS, $20\%$--$26\%$ lower FID, and $27\%$--$33\%$ lower FVD (similarly, under-fitting lowers SSIM from $0.7934\to 0.7742$ while improving PSNR by $+1.81$\,dB, LPIPS by $-30.6\%$, FID by $-36.1\%$, FVD by $-35.3\%$, and FBD by $-25.6\%$). Third, removing cross-camera seam jumps and splat floaters aligns multi-camera BEV representations consistently across all five families—cutting FBD from $340.6\to 261.5$ ($-23.2\%$) on chunk seams and $339.1\to 252.3$ ($-25.6\%$) on under-fitting, for five-family mean reductions of \textbf{$-17.5\%$ in FBD} ($317.5\to 262.0$), \textbf{$-34.7\%$ in FVD} ($257.1\to 167.9$), and \textbf{$-29.6\%$ in FID} ($4.37\to 3.08$).

\begin{table}[tbp]
  \centering
  \caption{\textbf{Quantitative 7-camera 3DGS restoration and BEV manifold alignment.} Evaluated over 7 cameras $\times$ 87 frames at native camera resolution using 8 Euler steps ($\sigma_{\text{src}}=0.3$). Each cell reports \emph{Degraded Input $\rightarrow$ Restored (Fixer)}, with bold marking the better of the two; FBD is measured in frozen BEVFormer~\citep{li2022bevformer} BEV feature space.}
  \label{tab:fixer}
  \scriptsize
  \setlength{\tabcolsep}{2.8pt}
  \begin{tabular*}{\linewidth}{@{\extracolsep{\fill}}l c c c c c c c@{}}
    \toprule
    \hd{3DGS Degradation} & \hd{PSNR (dB) $\uparrow$} & \hd{$\Delta$PSNR $\uparrow$} & \hd{SSIM $\uparrow$} & \hd{LPIPS $\downarrow$} & \hd{FID $\downarrow$} & \hd{FVD $\downarrow$} & \hd{FBD $\downarrow$} \\
    \midrule
    Chunk seams & $21.97 \rightarrow \mathbf{24.09}$ & $\mathbf{+2.12}$ & $0.7501 \rightarrow \mathbf{0.7759}$ & $0.4201 \rightarrow \mathbf{0.2961}$ & $5.13 \rightarrow \mathbf{3.18}$ & $280.3 \rightarrow \mathbf{152.1}$ & $340.6 \rightarrow \mathbf{261.5}$ \\
    Under-fitting & $21.95 \rightarrow \mathbf{23.76}$ & $+1.81$ & $\mathbf{0.7934} \rightarrow 0.7742$ & $0.4260 \rightarrow \mathbf{0.2957}$ & $4.76 \rightarrow \mathbf{3.04}$ & $261.4 \rightarrow \mathbf{169.2}$ & $339.1 \rightarrow \mathbf{252.3}$ \\
    Missing background & $23.28 \rightarrow \mathbf{24.35}$ & $+1.07$ & $\mathbf{0.7903} \rightarrow 0.7834$ & $0.3093 \rightarrow \mathbf{0.2781}$ & $3.61 \rightarrow \mathbf{2.88}$ & $230.5 \rightarrow \mathbf{167.9}$ & $292.2 \rightarrow \mathbf{251.7}$ \\
    Foreground noise & $24.78 \rightarrow \mathbf{25.46}$ & $+0.68$ & $\mathbf{0.8043} \rightarrow 0.7965$ & $0.2848 \rightarrow \mathbf{0.2534}$ & $4.10 \rightarrow \mathbf{3.14}$ & $250.8 \rightarrow \mathbf{172.8}$ & $310.5 \rightarrow \mathbf{265.4}$ \\
    Road noise & $24.94 \rightarrow \mathbf{25.56}$ & $+0.62$ & $\mathbf{0.8038} \rightarrow 0.7961$ & $0.2928 \rightarrow \mathbf{0.2507}$ & $4.27 \rightarrow \mathbf{3.15}$ & $262.7 \rightarrow \mathbf{177.3}$ & $305.1 \rightarrow \mathbf{278.9}$ \\
    \midrule
    \rowcolor{boschblue!7}\textbf{Mean across families} & $23.38 \rightarrow \mathbf{24.64}$ & $\mathbf{+1.26}$ & $\mathbf{0.7884} \rightarrow 0.7852$ & $0.3466 \rightarrow \mathbf{0.2748}$ & $4.37 \rightarrow \mathbf{3.08}$ & $257.1 \rightarrow \mathbf{167.9}$ & $317.5 \rightarrow \mathbf{262.0}$ \\
    \bottomrule
  \end{tabular*}
\end{table}

\needspace{5\baselineskip}
\section{Experiments}\label{sec:experiments}

\subsection{Data Foundation and Evaluation Protocol}\label{sec:setup}
\begin{figure}[tbp]
  \centering
  \includegraphics[width=\linewidth,height=0.45\textheight,keepaspectratio]{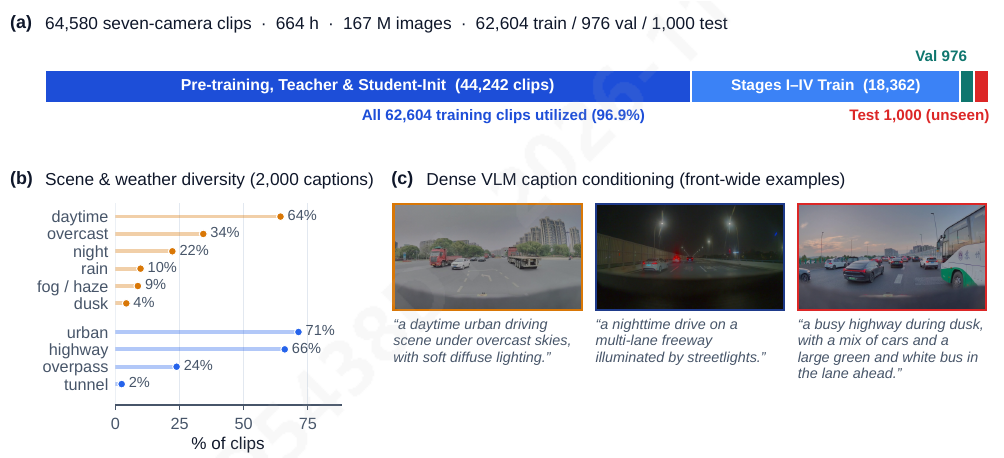}
  \caption{\textbf{Seven-camera driving corpus summary (64,580 clips).} (a) Total volume ($664$\,h, $167.4$\,M images) partitioned chronologically into $62{,}604$ training, $976$ validation, and $1{,}000$ test clips. (b) Multi-label illumination, weather, and road-topology tag distribution computed over 2,000 randomly sampled VLM captions (percentages sum above 100\%). (c) Representative clips with paired VLM captions.}
  \label{fig:dataset}
\end{figure}

\paragraph{Multi-camera driving corpus.}
BoschDreams\textsuperscript{7V} is trained and evaluated on a proprietary synchronized seven-camera driving corpus containing \textbf{64{,}580 clips} ($18$--$40$\,s each, totaling $664$\,hours and $23.9$\,M surround frames; $167.4$\,M camera images) downsampled to $1280\times 720$ at $10$\,Hz (from native vehicle camera feeds of $1920\times 1080$ on \texttt{FW}/\texttt{FT} and $1280\times 720$ on side/rear views) across diverse urban, highway, weather, and illumination conditions (\cref{fig:dataset}). Every surround frame provides LiDAR-fused 3D object tracklets, dense 3D static HDMap vectors, SLAM ego poses, and a structured vision--language scene caption. Clips are partitioned chronologically by recording timestamp to prevent geographic or temporal leakage: \textbf{1{,}000 held-out test clips} and \textbf{976 validation clips} are drawn from the most recent recordings, while all remaining \textbf{62{,}604 clips} are utilized for training—comprising a $44{,}242$-clip pre-training and teacher/student-initialization pool (Stages~I--II) and an $18{,}362$-clip curated post-training split (Stages~III--IV).

\paragraph{Evaluation protocol and metrics.}
Adapting the DrivingGen metric suite~\citep{drivinggen} to calibrated seven-camera surround rigs on our held-out driving corpus (\cref{sec:protocol,tab:protocol}), primary evaluations use a fixed 600-scene subset drawn from the $1{,}000$ held-out test clips at $10$\,Hz. We exclude conditioning frame $0$ and score frames $1$--$100$ ($10$\,s) at unified $832\times 480$ resolution (our ``720p'' model runs natively at $1280\times 704$ to align with the $16\times$ VAE+patch stride and is downsampled to $832\times 480$ for fair cross-resolution scoring). Single-view baselines are evaluated on the primary forward view (\texttt{FW}); native seven-view models are scored across all seven cameras ($4{,}200$ video streams per run) with equal-weight view averaging. We evaluate four complementary axes: (1)~\emph{video realism and coherence} via luminance exposure stability ($\text{Obj.\ q.}\uparrow$), CLIP-IQA+~\citep{wang2023clipiqa} perceptual quality ($\text{Subj.\ q.}\uparrow$), motion-sampled DINOv3~\citep{oquab2025dinov3} adjacent-frame similarity via SEA-RAFT~\citep{wang2024searaft} optical flow ($\text{V-cons.}\uparrow$), and 600-clip StyleGAN-V~\citep{skorokhodov2022styleganv} I3D Fr\'echet Video Distance~\citep{unterthiner2018fvd} ($\text{FVD}\downarrow$); (2)~\emph{dynamic agent fidelity} via tracked DINOv3 identity preservation ($\text{A-cons.}\uparrow$) and Cosmos-Reason1-7B~\citep{nvidia2025cosmosreason1} VLM-audited physical persistence without unnatural disappearance ($\text{A-pres.}\uparrow$); (3)~\emph{ego-motion accuracy} via UniDepthV2~\citep{piccinelli2025unidepthv2} monocular-SLAM-aligned Average Displacement Error ($\text{ADE}\downarrow$, meters), Dynamic Time Warping ($\text{DTW}\downarrow$), and Motion-Transformer~\citep{shi2022mtr} Fr\'echet Trajectory Distance ($\text{FTD}\downarrow$); and (4)~\emph{360\degree\ geometric consistency} via outdoor LoFTR~\citep{sun2021loftr} feature matches ($\text{CV match}\uparrow$) and confidence sums ($\text{CV conf.}\uparrow$) across all seven adjacent camera pairs. Recorded ground-truth videos (\emph{GT Measured Reference}) are audited under the identical pipeline as an empirical reference baseline rather than a theoretical upper bound: GT-only FVD is structurally inapplicable (\na), while annotation self-reference diagnostics yield $\text{ADE}=\text{DTW}=0.00$, $\text{FTD}\approx 2.62\times 10^{-9}$, trajectory quality $0.7469$, and trajectory consistency $0.5676$ over $532/600$ moving scenes (on a 30-clip GT video-SLAM control subset, the updated \texttt{FW} PnP pipeline achieves $0.92$\,m median / $3.08$\,m mean ADE; \cref{sec:trajectory}).

\subsection{Main Comparison with State-of-the-Art World Models}\label{sec:main}
\Cref{tab:main} compares public open-domain and interactive single-view world models, single-view Cosmos-Predict~2.5 (\emph{Cosmos~2.5~1V})~\citep{nvidia2025cosmos25}, our single-view HDMap-conditioned baseline (\emph{OmniDreams-1V$^\dagger$}) following OmniDreams-SV~\citep{nvidia2026omnidreams}, our 7-camera \emph{Cosmos-Transfer~2.5}~\citep{nvidia2025cosmos25} baseline, and \emph{BoschDreams\textsuperscript{7V}} in both its real-time 1-step ($480\text{p}$, $10.85$\,FPS) and high-fidelity 4-step ($720\text{p}$) configurations. Four empirical patterns govern \cref{tab:main}:
\begin{enumerate}
  \item \textbf{3D layout grounding suppresses unconditioned spatiotemporal drift:} Open-domain video generators and interactive camera/action world models that do not ingest structured 3D HDMap and bounding-box layouts (\emph{Causal Forcing}~\citep{zhu2026causalforcing}, \emph{LingBot-World}~\citep{team2026lingbot}, \emph{Wan2.2}~\citep{wan2025,wan2025wan22}, \emph{FastVideo Causal Wan2.2}~\citep{zhang2025fastvideo,zhang2025vsa}, \emph{HY-WorldPlay}~\citep{sun2025worldplay}, and \emph{Self Forcing}~\citep{huang2025selfforcing}; conditioned on the initial \texttt{FW} frame and scene prompt under the unified single-view protocol~\citep{drivinggen,horizondrive}) attain competitive single-frame perceptual scores ($\text{Subj.\ q.}\ 0.4262$--$0.5925$) yet suffer severe distribution divergence ($\text{FVD}\ 389.25$--$1177.90$) and trajectory error ($16.96$--$32.03$\,m ADE among models with $\ge 4$ valid SLAM scenes; \emph{Wan2.2} fails SLAM on $599/600$ scenes and is therefore omitted from trajectory aggregates in \cref{tab:main,tab:trajectory}). Conditioning on projected 3D HDMaps and bounding boxes cuts single-view FVD to $199.23$ (\emph{OmniDreams-1V$^\dagger$}) and surround FVD to \textbf{80.19} (1-step 480p) and \textbf{84.35} (4-step 720p) in \emph{BoschDreams\textsuperscript{7V}} ($74.93$ and $83.12$ on \texttt{FW}, with best surround views reaching $71.91$ and $73.80$).
  \item \textbf{Causal few-step memory vs.\ bidirectional multi-step sampling:} Relative to the 70-NFE (per 93-frame clip, $\approx 0.75$\,NFE/frame) \emph{Cosmos-Transfer~2.5} baseline ($106.99$ mean FVD, $14.50$\,m ADE, and $807.4$ DTW over $530/600$ valid scenes), \emph{BoschDreams\textsuperscript{7V}} reduces 7-view FVD by $25.1\%$ at 1-step 480p ($80.19$) and $21.2\%$ at 4-step 720p ($84.35$) while requiring only $0.25$\,NFE/frame ($2$ passes per 8-frame chunk, $\approx 3\times$ fewer NFE/frame) in the 1-step real-time mode. Comparing our two flagship tiers, 1-step 480p achieves lower FVD ($80.19$ vs.\ $84.35$ at 4-step 720p), whereas 4-step 720p substantially elevates perceptual sharpness ($\text{Subj.\ q.}\ 0.5645$, best view $0.5942$), temporal coherence ($\text{V-cons.}\ 0.8894$, best view $0.9163$), agent persistence ($\text{A-pres.}\ 0.6887$, best view $0.7961$), and cross-view correspondence ($380.50$ matches, second only to the 70-NFE bidirectional \emph{Cosmos-Transfer~2.5} at $415.67$).
  \item \textbf{Surround parallax and spatial resolution unlock SLAM localization:} Under the calibrated \texttt{FW} SLAM protocol (\cref{tab:trajectory}), \emph{BoschDreams\textsuperscript{7V}} achieves the lowest ego-motion error among all evaluated models at both resolutions: \textbf{7.02\,m ADE} (\textbf{411.6 DTW}, \textbf{65.39 FTD} over $544/600$ valid scenes) at 1-step $480\text{p}$ and \textbf{4.90\,m ADE} (\textbf{284.8 DTW}, \textbf{55.48 FTD} over $573/600$ valid scenes) at 4-step $720\text{p}$—approaching the $3.08$\,m real-video SLAM noise floor and outperforming single-view \emph{OmniDreams-1V$^\dagger$} ($13.49$\,m ADE) because 14-directed-edge surround coupling and layout-anchored horizon geometry reduce inter-chunk yaw drift.
  \item \textbf{Why generated streams can exceed the GT Measured Reference on proxy metrics:} Recorded fleet cameras experience physical auto-exposure stepping under bridges and natural lead-vehicle occlusions ($\text{Obj.\ q.}=0.7624$, $\text{A-pres.}=0.6669$). Because our causal generator conditions on continuous 3D tracklets and light-noise KV memory ($\sigma_{\text{write}}=0.13$), \emph{BoschDreams\textsuperscript{7V}} (4-step 720p) surpasses the GT Measured Reference on exposure stability ($0.9001$), CLIP-IQA+ ($0.5645$ vs.\ $0.5402$), temporal consistency ($0.8894$ vs.\ $0.8851$), and physical agent persistence ($0.6887$ vs.\ $0.6669$), while recovering $76.0\%$ of GT cross-view LoFTR matches ($380.50$ vs.\ $500.62$; best pair $510.62$).
\end{enumerate}

\begin{table}[tbp]
  \centering
  \caption{\textbf{Comparison with state-of-the-art single-view and surround-view world models (600-scene benchmark, frames 1--100 at 10\,Hz).} Single-view models are scored on \texttt{FW} (cross-view LoFTR inapplicable, \na); seven-view models report unweighted 7V mean and best surround view. \emph{Wan2.2} is omitted from trajectory columns ($<4$ valid SLAM scenes; \cref{tab:trajectory}). Bold: best; underline: second best.}
  \label{tab:main}
  \scriptsize
  \setlength{\tabcolsep}{1.5pt}
  \begin{tabular*}{\linewidth}{@{\extracolsep{\fill}}lc cccc cc ccc cc@{}}
    \toprule
    & & \multicolumn{4}{c}{\hd{Video Quality \& Realism}} & \multicolumn{2}{c}{\hd{Agent Plausibility}} & \multicolumn{3}{c}{\hd{Ego Trajectory}} & \multicolumn{2}{c}{\hd{Cross-View}} \\
    \cmidrule(lr){3-6}\cmidrule(lr){7-8}\cmidrule(lr){9-11}\cmidrule(l){12-13}
    \hd{Method} & \hd{Views} & \hd{Obj.q.$\uparrow$} & \hd{Subj.q.$\uparrow$} & \hd{V-cons.$\uparrow$} & \hd{FVD$\downarrow$} & \hd{A-cons.$\uparrow$} & \hd{A-pres.$\uparrow$} & \hd{ADE$\downarrow$} & \hd{DTW$\downarrow$} & \hd{FTD$\downarrow$} & \hd{Match$\uparrow$} & \hd{Conf.$\uparrow$} \\
    \midrule
    \rowcolor{boschgray!8}\textit{GT Measured Reference} & 7V & \textit{0.7624} & \textit{0.5402} & \textit{0.8851} & \na & \textit{0.7794} & \textit{0.6669} & \textit{0.00} & \textit{0.0} & \textit{0.00} & \textit{500.62} & \textit{272.84} \\
    \midrule
    Causal Forcing~\citep{zhu2026causalforcing} & 1V & 0.6800 & 0.4295 & 0.7367 & 650.71 & 0.5854 & 0.4385 & 20.70 & 1870.5 & 89.09 & \na & \na \\
    LingBot-World~\citep{team2026lingbot} & 1V & 0.6490 & 0.4262 & 0.7373 & 389.25 & 0.6429 & 0.5371 & 16.96 & 1218.5 & 134.81 & \na & \na \\
    Wan2.2~\citep{wan2025,wan2025wan22} & 1V & \best{0.9180} & 0.4801 & 0.8124 & 713.69 & 0.6131 & 0.6034 & \na & \na & \na & \na & \na \\
    FastVideo Causal Wan2.2~\citep{zhang2025fastvideo,zhang2025vsa} & 1V & 0.7833 & \best{0.5925} & \best{0.8953} & 1177.90 & \best{0.7579} & 0.4992 & 23.09 & 1619.9 & 88.89 & \na & \na \\
    HY-WorldPlay~\citep{sun2025worldplay} & 1V & 0.8850 & 0.4910 & 0.8058 & 769.15 & \secbest{0.7570} & 0.5307 & 32.03 & 2762.4 & 86.63 & \na & \na \\
    Self Forcing~\citep{huang2025selfforcing} & 1V & 0.8150 & \secbest{0.5657} & 0.7974 & 1056.56 & 0.6860 & 0.5617 & 24.90 & 1967.6 & 153.05 & \na & \na \\
    Cosmos-Transfer~2.5~\citep{nvidia2025cosmos25} & 7V & 0.7667 & 0.5041 & 0.8613 & 106.99 & 0.7120 & 0.6491 & 14.50 & 807.4 & 148.41 & \best{415.67} & \best{209.69} \\
    \subrow{\quad\textit{best surround view}} & \subrow{1V} & \subrow{0.8619} & \subrow{0.5547} & \subrow{0.8982} & \subrow{100.04} & \subrow{0.7378} & \subrow{0.7935} & \subrow{\na} & \subrow{\na} & \subrow{\na} & \subrow{653.67} & \subrow{373.62} \\
    Cosmos~2.5~1V~\citep{nvidia2025cosmos25} & 1V & 0.8700 & 0.4294 & 0.8256 & 260.97 & 0.7295 & 0.6296 & 20.13 & 1427.1 & 106.81 & \na & \na \\
    OmniDreams-1V$^\dagger$~\citep{nvidia2026omnidreams} & 1V & 0.7883 & 0.4377 & 0.7931 & 199.23 & 0.7132 & 0.6296 & 13.49 & 791.6 & 80.42 & \na & \na \\
    \midrule
    \rowcolor{boschblue!7}\textbf{BoschDreams\textsuperscript{7V} (1-step)} & \textbf{7V} & 0.8510 & 0.4937 & 0.8494 & \best{80.19} & 0.6820 & \secbest{0.6505} & \secbest{7.02} & \secbest{411.6} & \secbest{65.39} & 307.78 & 126.10 \\
    \rowcolor{boschblue!4}\subrow{\quad\textit{best surround view}} & \subrow{1V} & \subrow{0.9183} & \subrow{0.5286} & \subrow{0.8946} & \subrow{71.91} & \subrow{0.7116} & \subrow{0.7375} & \subrow{\na} & \subrow{\na} & \subrow{\na} & \subrow{410.36} & \subrow{187.68} \\
    \rowcolor{boschblue!9}\textbf{BoschDreams\textsuperscript{7V} (4-step)} & \textbf{7V} & \secbest{0.9001} & 0.5645 & \secbest{0.8894} & \secbest{84.35} & 0.7359 & \best{0.6887} & \best{4.90} & \best{284.8} & \best{55.48} & \secbest{380.50} & \secbest{183.25} \\
    \rowcolor{boschblue!4}\subrow{\quad\textit{best surround view}} & \subrow{1V} & \subrow{0.9290} & \subrow{0.5942} & \subrow{0.9163} & \subrow{73.80} & \subrow{0.7594} & \subrow{0.7961} & \subrow{\na} & \subrow{\na} & \subrow{\na} & \subrow{510.62} & \subrow{288.41} \\
    \bottomrule
  \end{tabular*}
  \tablenote{$^\dagger$In-house single-view HDMap-conditioned reimplementation following OmniDreams-SV~\citep{nvidia2026omnidreams}.}
\end{table}

\subsection{Long-Horizon Rollout Stability}\label{sec:longhorizon}
To quantify autoregressive degradation past the initial $10$\,s prefix, \cref{tab:longhorizon} pairs early rollout windows (\texttt{First 100} frames, $0$--$10$\,s) against late windows (\texttt{Last 100} frames, $10.4$--$20.4$\,s on 205-frame native 7V rollouts; $12.4$--$22.4$\,s on \emph{Cosmos-Transfer~2.5}). Three structural behaviors stand out. First, public single-view models drift noticeably over time: \emph{Wan2.2} loses $-0.05$ in temporal consistency ($0.81\to 0.76$) at $726.24$ tail FVD, \emph{LingBot-World} rises to $463.91$ tail FVD while dropping $-0.05$ in exposure stability ($0.65\to 0.60$), and \emph{Cosmos~2.5~1V} rises from $260.97\to 338.06$ FVD, whereas layout-conditioned \emph{OmniDreams-1V} remains comparatively bounded ($199.23\to 229.42$ FVD). Second, because \emph{Cosmos-Transfer~2.5} re-conditions each chunk without causal KV memory, errors compound across chunk boundaries, causing a $4.03\times$ FVD surge ($106.99\to 430.91$) and a $16.5\%$ drop in cross-view LoFTR matches ($415.67\to 346.95$). Third, our Stage~IV curriculum progressively stabilizes long-horizon rollouts. Whereas \emph{Stage~III Self-Forcing} suffers a $2.91\times$ tail FVD surge ($128.44\to 373.58$) and loses $21.6\%$ of cross-camera matches ($289.82\to 227.13$), \emph{Stage~IV LAK} cuts late-window 7V FVD by $66.6\%$ ($373.58\to 124.91$) while retaining $86.7\%$ of surround correspondences ($318.63\to 276.41$). Across our three \emph{BoschDreams\textsuperscript{7V}} operating points, the real-time $480\text{p}$ 1-step model achieves \textbf{119.53} late-window 7V FVD ($-68.0\%$ vs.\ Stage~III) with flat $0.49$ perceptual quality and $0.85$ temporal consistency across both windows. The $480\text{p}$ 4-step model pushes late-window 7V FVD down further to \textbf{102.17} ($-72.7\%$ vs.\ Stage~III, with only $+11.0\%$ early-to-late drift), reaches $0.87$ late-window temporal consistency, and retains $94.9\%$ of cross-view matches ($326.28\to 309.65$, confidence $177.52\to 149.93$). The flagship $720\text{p}$ 4-step model achieves the lowest late-window 7V FVD (\textbf{99.67}, $-73.3\%$ vs.\ Stage~III) alongside the strongest late-window subjective quality ($0.53$), temporal consistency ($0.88$), agent fidelity ($0.75 / 0.75$), and cross-view matching ($330.71$, confidence $165.74$).

\begin{table}[tbp]
  \centering
  \caption{\textbf{Long-horizon stability across early (\texttt{First 100}) and late (\texttt{Last 100}) 100-frame rollout windows.} Bold marks the best and underline the second best in each window across all 600-scene models (GT-only FVD is inapplicable, \na).}
  \label{tab:longhorizon}
  \fontsize{6.6pt}{7.8pt}\selectfont
  \setlength{\tabcolsep}{1.2pt}
  \begin{tabular*}{\linewidth}{@{\extracolsep{\fill}}llc cccc cc cc@{}}
    \toprule
    \hd{Method} & \hd{Window} & \hd{Views} & \hd{Obj.q.$\uparrow$} & \hd{Subj.q.$\uparrow$} & \hd{V-cons.$\uparrow$} & \hd{FVD$\downarrow$} & \hd{A-cons.$\uparrow$} & \hd{A-pres.$\uparrow$} & \hd{Match$\uparrow$} & \hd{Conf.$\uparrow$} \\
    \midrule
    \rowcolor{boschgray!8}\textit{GT Measured Reference} & First 100 & 7V & \textit{0.76} & \textit{0.54} & \textit{0.89} & \na & \textit{0.78} & \textit{0.67} & \textit{500.62} & \textit{272.84} \\
    \midrule
    LingBot-World~\citep{team2026lingbot} & First 100 / Last 100 & 1V & 0.65 / 0.60 & 0.43 / 0.43 & 0.74 / 0.73 & 389.25 / 463.91 & 0.64 / 0.66 & 0.54 / 0.62 & \na & \na \\
    Wan2.2~\citep{wan2025,wan2025wan22} & First 100 / Last 100 & 1V & \best{0.92} / \best{0.92} & 0.48 / 0.48 & 0.81 / 0.76 & 713.69 / 726.24 & 0.61 / 0.64 & 0.60 / 0.58 & \na & \na \\
    Cosmos~2.5~1V~\citep{nvidia2025cosmos25} & First 100 / Last 100 & 1V & 0.87 / 0.80 & 0.43 / 0.43 & 0.83 / 0.83 & 260.97 / 338.06 & \secbest{0.73} / 0.71 & 0.63 / 0.59 & \na & \na \\
    OmniDreams-1V~\citep{nvidia2026omnidreams} & First 100 / Last 100 & 1V & 0.79 / 0.79 & 0.44 / 0.44 & 0.79 / 0.79 & 199.23 / 229.42 & 0.71 / 0.72 & 0.63 / 0.61 & \na & \na \\
    \midrule
    Cosmos-Transfer~2.5~\citep{nvidia2025cosmos25} & First 100 / Last 100 & 7V & 0.77 / 0.80 & \secbest{0.50} / 0.47 & \secbest{0.86} / 0.84 & 106.99 / 430.91 & 0.71 / \secbest{0.73} & 0.65 / \best{0.75} & \best{415.67} / \best{346.95} & \best{209.69} / \secbest{165.73} \\
    \midrule
    Stage~III Self-Forcing & First 100 / Last 100 & 7V & 0.89 / 0.87 & 0.47 / 0.42 & 0.84 / 0.82 & 128.44 / 373.58 & 0.67 / 0.71 & 0.63 / \secbest{0.73} & 289.82 / 227.13 & 117.86 / 82.30 \\
    Stage~IV LAK & First 100 / Last 100 & 7V & \best{0.92} / 0.85 & 0.49 / 0.48 & \secbest{0.86} / 0.85 & \secbest{80.97} / 124.91 & 0.69 / 0.72 & 0.65 / 0.70 & 318.63 / 276.41 & 131.87 / 105.04 \\
    \rowcolor{boschblue!5}\textbf{BoschDreams\textsuperscript{7V} (480p, 1s)} & First 100 / Last 100 & \textbf{7V} & 0.85 / 0.81 & 0.49 / \secbest{0.49} & 0.85 / 0.85 & \best{80.19} / 119.53 & 0.68 / 0.70 & 0.65 / 0.70 & 307.78 / 273.22 & 126.10 / 102.96 \\
    \rowcolor{boschblue!7}\textbf{BoschDreams\textsuperscript{7V} (480p, 4s)} & First 100 / Last 100 & \textbf{7V} & 0.87 / 0.88 & 0.49 / 0.47 & \secbest{0.86} / \secbest{0.87} & 92.02 / \secbest{102.17} & 0.69 / 0.68 & \secbest{0.66} / 0.67 & 326.28 / 309.65 & 177.52 / 149.93 \\
    \rowcolor{boschblue!9}\textbf{BoschDreams\textsuperscript{7V} (720p, 4s)} & First 100 / Last 100 & \textbf{7V} & \secbest{0.90} / \secbest{0.89} & \best{0.56} / \best{0.53} & \best{0.89} / \best{0.88} & 84.35 / \best{99.67} & \best{0.74} / \best{0.75} & \best{0.69} / \best{0.75} & \secbest{380.50} / \secbest{330.71} & \secbest{183.25} / \best{165.74} \\
    \bottomrule
  \end{tabular*}
\end{table}

\begin{figure}[tbp]
  \centering
  \includegraphics[width=\linewidth,height=0.38\textheight,keepaspectratio]{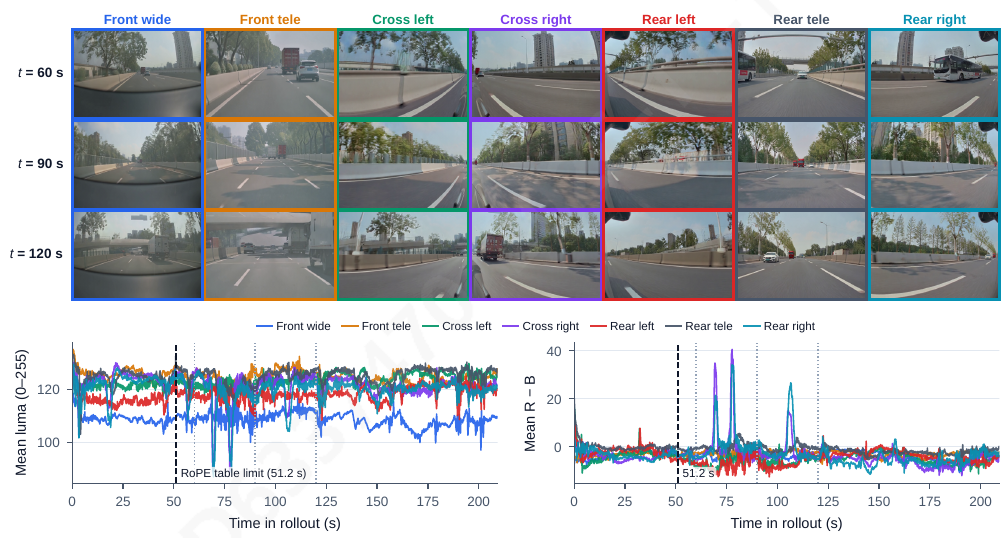}
  \caption{\textbf{Continuous 209\,s (2,093 frames per camera; 14,651 total camera images) surround-view rollout across exact rotary position rebases.} Comprises 1 initial conditioning frame, 261 full 8-frame chunks, and 4 trailing frames ($209.3$\,s at $10$\,Hz). Top: seven-camera keyframes at $60, 90,$ and $120$\,s with camera border colors matching the traces below. Bottom: per-camera luminance and $R{-}B$ color balance past the $51.2$\,s RoPE table limit (dashed line).}
  \label{fig:longhorizon}
\end{figure}

\Cref{fig:longhorizon} further stress-tests continuous streaming over a \textbf{209\,s (2{,}093-frame per camera: 1 initial frame + 261 chunks $\times$ 8 frames + 4 trailing frames, totaling 14{,}651 images across 7 cameras)} rollout of the 720p model—exceeding the $51.2$\,s pre-tabulated RoPE horizon by more than $4\times$. Across multiple exact rotary rebases, all seven cameras remain geometrically locked and photometrically stable, with mean luminance and red--blue color balance shifting by less than $3.8$ intensity levels between the opening and closing $20$\,s.

\subsection{Ablation Studies}\label{sec:ablation_results}
\Cref{tab:ablation} isolates our core curriculum, supervision, and inference-hyperparameter choices across two structured parts on the full 600-scene benchmark.

\paragraph{Part I: Four-stage curriculum, Layout-Aware Kernel Score (LAK), and Antiphon Forcing.}
Rows~(1--5) trace the progression from offline teacher to real-time memory-aware generator at matched 480p resolution (4 steps in Stages~III--IV).
First, switching directly from the 70-NFE bidirectional Stage~I teacher ($0.5806$ CLIP-IQA+, $82.12$ 7V FVD, $382.40$ LoFTR matches at $0.14$\,FPS) to 35-step causal streaming in Stage~II ($0.16$\,FPS) exposes severe autoregressive compounding error ($135.61$ 7V FVD, $152.91$ \texttt{FW} FVD, $16.41$\,m ADE).
Second, Stage~III Self-Forcing unrolls few-step generation on self-generated KV history, accelerating inference by $31.7\times$ ($0.16\to 5.07$\,FPS eager; $6.63$\,FPS with \texttt{compile+FP8}) and cutting ego-trajectory ADE from $16.41\to 10.27$\,m ($489/600$ valid scenes), though frozen-teacher score matching leaves residual blur ($128.44$ 7V FVD, $289.82$ matches).
Third, adding \emph{Stage~IV LAK} with a shared rank-32 LoRA (row~4) anchors multi-encoder perceptual and layout features to real recordings, reducing 7V FVD by $37.0\%$ ($128.44\to 80.97$), cutting ADE by $47.5\%$ ($10.27\to 5.39$\,m over $561/600$ scenes), and lifting matches to $318.63$.
Finally, introducing dual-role $\Delta W_{\mathcal{W}},\Delta W_{\mathcal{D}}$ decoupling and depth-2 causal-VAE decoder-VJP memory backpropagation (row~5, full \emph{Antiphon Forcing}) unlocks substantial long-horizon, geometric, and forward-view gains at matched 480p 4-step inference (\cref{tab:longhorizon,tab:ablation,tab:pervview_ablation,tab:trajectory}). On the \texttt{First 100} window, Antiphon Forcing cuts \texttt{FW} FVD by $12.0\%$ ($95.02\to 83.59$) and \texttt{FT} FVD by $10.5\%$ ($113.01\to 101.12$), improves temporal consistency ($0.8559\to 0.8572$), agent consistency ($0.6886\to 0.6944$), agent persistence ($0.6511\to 0.6635$), cross-view matches ($318.63\to 326.28$), and SLAM ego-trajectory accuracy ($5.39\to \mathbf{5.17}$\,m ADE; DTW $367.13\to 300.74$, $-18.1\%$; valid scenes $561\to 571/600$). Most critically, on the matched 480p 4-step \texttt{Last 100} tail window (\cref{tab:longhorizon}), full Antiphon Forcing outperforms shared-LoRA LAK decisively—reducing late-window 7V FVD by \textbf{18.2\%} ($124.91\to \mathbf{102.17}$), raising late-window temporal consistency ($0.85\to 0.87$), and boosting late-window cross-view matches by $+12.0\%$ ($276.41\to 309.65$) and confidence by $+42.7\%$ ($105.04\to 149.93$). In the short \texttt{First 100} window, 7V mean FVD rises slightly ($80.97\to 92.02$) solely in side/rear views (\cref{tab:pervview_ablation}) as memory BPTT prioritizes temporal stability and forward-horizon geometry over peripheral micro-texture variance.

\paragraph{Part II: Inference denoising steps and spatial resolution.}
Within the 480p Stage~IV Antiphon Forcing checkpoint, varying $N_{\text{step}}$ from $1\to 2\to 4$ exposes a clear speed--geometry trade-off (\cref{tab:ablation,tab:pervview_all,tab:pervview_ablation}). Single-step inference achieves the lowest short-horizon FVD ($80.19$ 7V, $74.93$ \texttt{FW}) and real-time throughput ($10.85$\,FPS, $7.02$\,m ADE). Increasing $N_{\text{step}}$ to $2$ and $4$ monotonically strengthens geometric structure and long-horizon stability—improving temporal consistency ($0.8494\to 0.8523\to 0.8572$), agent consistency ($0.6820\to 0.6859\to 0.6944$), cross-camera LoFTR matches ($307.78\to 310.11\to 326.28$), SLAM accuracy ($7.02\to 5.66\to 5.17$\,m ADE; DTW $411.56\to 353.07\to 300.74$; valid scenes $544\to 552\to 571/600$), and late-window 7V FVD ($119.53\to 102.17$ in \cref{tab:longhorizon})—while slightly raising short-horizon 7V FVD ($80.19\to 80.93\to 92.02$) as repeated deterministic denoising damps high-frequency stochastic micro-texture. Scaling spatial resolution to $720\text{p}$ ($1280\times 704$, $N_{\text{step}}=4$, FP8 attention at $2.03$\,FPS) resolves this tension across every axis: perceptual quality rises to \textbf{0.5645}, temporal consistency to \textbf{0.8894}, 7V/\texttt{FW} FVD drops back to \textbf{84.35 / 83.12} (\texttt{Last 100} \textbf{99.67}), agent consistency/persistence reaches \textbf{0.7359 / 0.6887}, cross-view matches reach \textbf{380.50} ($76.0\%$ of GT), and ego-trajectory error drops to \textbf{4.90\,m ADE} (\textbf{284.8 DTW}, \textbf{55.48 FTD}; improving across 1-, 2-, and 4-step 720p from $6.10\to 5.15\to 4.90$\,m in \cref{tab:trajectory}).

\begin{table}[tbp]
  \centering
  \caption{\textbf{Ablation study of BoschDreams\textsuperscript{7V} across all 600 test scenes.} \emph{Part~I} isolates each curriculum stage, \emph{Layout-Aware Kernel Score (LAK)}, and full \emph{Antiphon Forcing} at matched 480p 4-step inference (Stages~III--IV). \emph{Part~II} ablates inference steps ($N_{\text{step}}\in\{1,2,4\}$) and spatial resolution ($480\text{p}$ vs.\ $720\text{p}$). Within each part, bold marks the best and underline the second best.}
  \label{tab:ablation}
  \fontsize{7.6pt}{8.8pt}\selectfont
  \setlength{\tabcolsep}{1.0pt}
  \begin{tabular*}{\linewidth}{@{\extracolsep{\fill}}l cccc cc cc c@{}}
    \toprule
    \hd{Variant / Configuration} & \hd{FPS$\uparrow$} & \hd{Subj.q.$\uparrow$} & \hd{V-cons.$\uparrow$} & \hd{7V FVD$\downarrow$} & \hd{FW FVD$\downarrow$} & \hd{A-cons.$\uparrow$} & \hd{A-pres.$\uparrow$} & \hd{Match$\uparrow$} & \hd{ADE$\downarrow$} \\
    \midrule
    \multicolumn{10}{@{}l}{\textbf{Part I: Four-Stage Curriculum \& Antiphon Forcing} \textit{(480p, 4 steps in Stages III--IV; 70 NFE in Stages I--II)}} \\
    (1) Stage~I Bidirectional Teacher (35-step, CFG 5) & 0.14 & \best{0.5806} & 0.8409 & \secbest{82.12} & \secbest{86.92} & 0.6884 & \best{0.6793} & \best{382.40} & 7.56 \\
    (2) Stage~II Causal Student (35-step, CFG 5) & 0.16 & 0.4522 & 0.8291 & 135.61 & 152.91 & 0.6359 & 0.5992 & 310.05 & 16.41 \\
    (3) Stage~III Self-Forcing (rollout DMD only) & \best{6.63} & 0.4724 & 0.8408 & 128.44 & 138.75 & 0.6703 & 0.6280 & 289.82 & 10.27 \\
    (4) + Stage~IV LAK (shared LoRA, no mem.\ VJP) & \best{6.63} & \secbest{0.4941} & \secbest{0.8559} & \best{80.97} & 95.02 & \secbest{0.6886} & 0.6511 & 318.63 & \secbest{5.39} \\
    (5) + \textbf{Antiphon Forcing} (LAK + Dual-LoRA + Mem.\ VJP) & \best{6.63} & 0.4892 & \best{0.8572} & 92.02 & \best{83.59} & \best{0.6944} & \secbest{0.6635} & \secbest{326.28} & \best{5.17} \\
    \midrule
    \multicolumn{10}{@{}l}{\textbf{Part II: Denoising Steps \& Spatial Resolution} \textit{(Full Stage~IV Antiphon Forcing)}} \\
    \rowcolor{boschblue!7}480p ($832\times 480$), $N_{\text{step}}=1$, compile+FP8 \textbf{(real-time)} & \best{10.85} & \secbest{0.4937} & 0.8494 & \best{80.19} & \best{74.93} & 0.6820 & 0.6505 & 307.78 & 7.02 \\
    480p ($832\times 480$), $N_{\text{step}}=2$, compile+FP8 & \secbest{8.62} & \secbest{0.4937} & 0.8523 & \secbest{80.93} & \secbest{78.12} & 0.6859 & \secbest{0.6686} & 310.11 & 5.66 \\
    480p ($832\times 480$), $N_{\text{step}}=4$, compile+FP8 & 6.63 & 0.4892 & \secbest{0.8572} & 92.02 & 83.59 & \secbest{0.6944} & 0.6635 & \secbest{326.28} & \secbest{5.17} \\
    \rowcolor{boschblue!9}720p ($1280\times 704$), $N_{\text{step}}=4$, FP8 attn.\ \textbf{(high-res)} & 2.03 & \best{0.5645} & \best{0.8894} & 84.35 & 83.12 & \best{0.7359} & \best{0.6887} & \best{380.50} & \best{4.90} \\
    \bottomrule
  \end{tabular*}
\end{table}

\subsection{Per-Camera Surround Fidelity and Cross-View Geometric Analysis}\label{sec:perview}
\begin{figure}[tbp]
  \centering
  \includegraphics[width=0.98\linewidth,height=0.38\textheight,keepaspectratio]{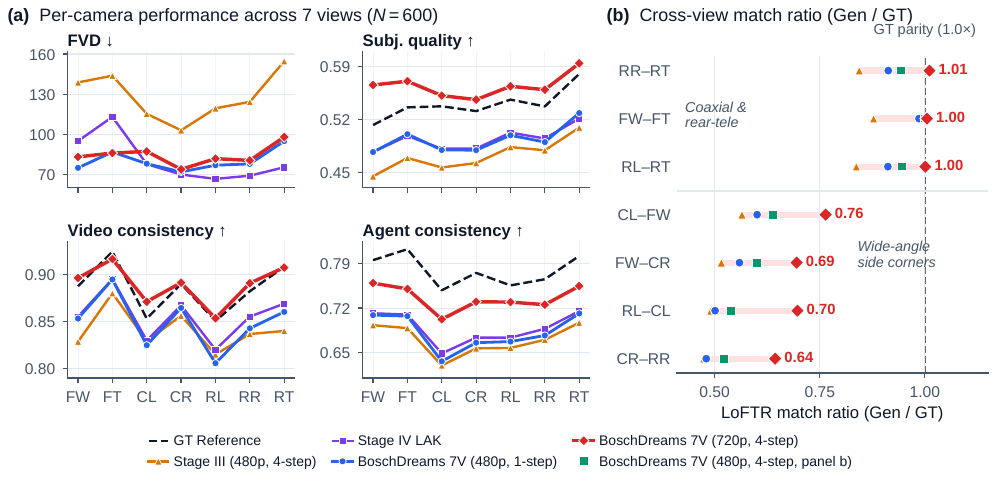}
  \caption{\textbf{Per-camera metrics across seven views and cross-camera LoFTR match ratio across seven rig edges ($N=600$).} (a) Per-camera FVD ($\downarrow$), CLIP-IQA+ subjective quality ($\uparrow$), temporal consistency ($\uparrow$), and agent consistency ($\uparrow$) across \texttt{FW, FT, CL, CR, RL, RR, RT}. (b) Generated-to-GT LoFTR match ratio ($\uparrow$) across the seven physical rig edges.}
  \label{fig:perview}
\end{figure}

Aggregating seven cameras into a single scalar hides systematic optical differences across a heterogeneous driving rig. \Cref{fig:perview} dissects performance across all seven individual views (a) and all seven calibrated adjacent camera pairs (b; per-camera numerical tables in \cref{sec:pervview,tab:pervview_all,tab:pervview_ablation}). Three physical patterns emerge:
\begin{enumerate}
  \item \textbf{Uniform 360\degree\ FVD and quality gains across wide and tele optics:} Every camera of BoschDreams\textsuperscript{7V} improves substantially over Stage~III Self-Forcing ($30\%$--$46\%$ per-view FVD reduction at 1-step 480p; \cref{fig:perview}a, \cref{tab:pervview_all,tab:pervview_ablation}), with all seven individual views ($71.91$--$94.90$ at 1-step 480p and $73.80$--$97.99$ at 4-step 720p) achieving lower FVD than the best single surround view of Cosmos-Transfer~2.5 ($100.04$; \cref{tab:main}). Under 480p 1-step, narrow-FOV telephoto views (\texttt{FT} $30^\circ$, \texttt{RT} $30^\circ$) exhibit higher FVD ($86.77, 94.90$) due to magnified optical flow; at 720p 4-step, both telephoto views achieve the highest perceptual quality ($0.5706$--$0.5942$, exceeding the GT Measured Reference across all seven views) and temporal consistency ($0.9073$--$0.9163$, tracking the $0.9081$--$0.9242$ GT profile).
  \item \textbf{Consistent multi-camera agent identity and blind-spot persistence:} Tracked agent appearance consistency (\cref{fig:perview}a, bottom right) rises uniformly from Stage~III ($0.6296$--$0.6970$) to Stage~IV LAK ($0.6487$--$0.7153$) and 720p BoschDreams\textsuperscript{7V} ($0.7024$--$0.7594$), while physical persistence peaks on blind-spot cameras \texttt{RL} and \texttt{RR} ($0.7450$--$0.7961$; \cref{tab:pervview_all}), matching recorded GT ($0.7335$--$0.7662$) as overtaking vehicles traverse side frustums smoothly.
  \item \textbf{Coaxial/rear-telephoto vs.\ wide-angle corner correspondence:} On the coaxial front pair (\mbox{FW--FT}) and rear-telephoto pairs (\mbox{RR--RT}, \mbox{RL--RT}), BoschDreams\textsuperscript{7V} recovers \textbf{91\%--99\%} of ground-truth LoFTR correspondences at 1-step 480p and reaches \textbf{100\%--101\%} GT parity at 4-step 720p (\cref{fig:perview}b). On wide-parallax $50^\circ$--$80^\circ$ yaw transitions (\mbox{CL--FW}, \mbox{FW--CR}, \mbox{RL--CL}, \mbox{CR--RR}), match recovery climbs from $47\%$--$57\%$ in Stage~III to \textbf{64\%--76\%} at 720p, pinpointing wide-baseline perspective foreshortening as the primary remaining frontier for 3D-consistent synthesis.
\end{enumerate}

\subsection{Limitations and Future Work}\label{sec:limitations}
Despite real-time seven-camera streaming ($10.85$\,FPS) at 480p, three practical boundaries remain:
\begin{itemize}
  \item \textbf{High-resolution throughput:} Native $720\text{p}$ runs at $2.03$\,FPS (4-step), bounded by $3{,}520$ tokens/frame and the $\sigma_{\text{write}}=0.13$ write pass; FP4 quantization and sparse KV eviction are natural next levers.
  \item \textbf{Wide-baseline foreshortening:} Coaxial/rear-tele pairs reach $100\%$--$101\%$ GT LoFTR parity at 720p (\cref{fig:perview}b), whereas $50^\circ$--$80^\circ$ seams recover $64\%$--$76\%$ during close pass-bys, motivating explicit 3D ray/Pl\"ucker conditioning.
  \item \textbf{Rare physical phenomena:} Extreme fluid-optical effects (heavy tire spray, wet specular glare, multi-body collisions) remain challenging for latent flow models.
\end{itemize}

\paragraph{Responsible deployment.}
BoschDreams\textsuperscript{7V} serves as a SIL/HIL simulation multiplier that complements rather than replaces on-road validation; policies trained on synthetic streams undergo final regression gates on held-out fleet logs.

\paragraph{Conclusion and future directions.}
We presented \textbf{BoschDreams\textsuperscript{7V}}, a layout-conditioned surround-view world model unifying 360\degree\ coherence, real-time streaming ($10.85$\,FPS at 1-step 480p), multi-minute stability, 3DGS restoration, and AI-agent orchestration. On our 600-scene benchmark, our four-stage curriculum, LAK, and \emph{Antiphon Forcing} achieve $80.19$ 7V FVD ($74.93$ \texttt{FW}, $7.02$\,m ADE) at 1-step 480p and $84.35$ 7V FVD, $380.50$ LoFTR matches, and $4.90$\,m ADE at 4-step 720p. Looking ahead, we advance five directions:
\begin{enumerate}
  \item \textbf{Prompt-driven world editing:} Zero-shot synthesis of rare weather, lighting, and hazards from VLM specs.
  \item \textbf{Unified multimodal generation:} Co-generating calibrated LiDAR, 4D radar, occupancy, and seven cameras from one latent state.
  \item \textbf{Closed-loop training:} Embedding vision-action policies for interactive RL and safety-critical benchmarking.
  \item \textbf{Recursive Self-Improvement (RSI):} Coupling corner-case rollouts with 3D/4D auto-labeling~\citep{jiang20243dsflabelling} to iteratively upgrade onboard stacks.
  \item \textbf{Autonomous AI-agent iteration:} Scaling our ticket-to-world agent (\cref{sec:closedloop}) into a self-directed adversarial failure miner.
\end{enumerate}
\needspace{4\baselineskip}
\section*{Acknowledgements}\label{sec:acknowledgements}
We thank the Bosch (CN) data collection and fleet-annotation teams for the synchronized seven-camera driving corpus, LiDAR-fused 3D tracklets, and HDMap ground truth. We are grateful to our infrastructure colleagues for supporting large-scale distributed training, low-latency streaming deployment, and the BEVFormer/DrivingGen evaluation pipelines, and to the planners and perception teams for closed-loop validation feedback.

\clearpage
\bibliographystyle{plainnat}
\bibliography{references}

\clearpage
\appendix
\renewcommand*{\thesection}{\Alph{section}}
\setcounter{section}{0}
\section*{Supplementary Material}\addcontentsline{toc}{section}{Supplementary Material}
\markboth{Supplementary Material}{Supplementary Material}

This supplementary material provides reference tables for abbreviations and mathematical notation (\cref{sec:abbrev}), full hardware, model, and closed-loop simulation configurations (\cref{sec:config}), and complete evaluation protocols, ego-trajectory eligibility counts, and per-camera surround breakdowns across all seven calibrated views (\cref{sec:protocol}).

\section{Abbreviations and Notation}\label{sec:abbrev}
\Cref{tab:abbrev,tab:notation} collect all acronyms and mathematical symbols used throughout the main report, aligning architectural dimensions, curriculum objectives, streaming KV-cache indices, and Diffusion Fixer notation.

\begin{table}[bp]
  \centering
  \caption{\textbf{Abbreviations used throughout this technical report.}}
  \label{tab:abbrev}
  \scriptsize
  \setlength{\tabcolsep}{3.0pt}
  \begin{tabularx}{\linewidth}{@{}>{\bfseries}l>{\RaggedRight}X>{\bfseries}l>{\RaggedRight}X>{\bfseries}l>{\RaggedRight}X@{}}
    \toprule
    3DGS & 3D Gaussian Splatting & DTW & Dynamic Time Warping & MMD & Maximum Mean Discrepancy \\
    7V / 1V & Seven-View / Single-View & EMA & Exponential Moving Average & NFE & Number of Function Evaluations \\
    A-cons. & Agent Appearance Consistency & FOV & Field of View & Obj. q. & Objective Flicker Stability \\
    A-pres. & Agent Physical Persistence & FPS & Per-Camera Frames Per Second & PPU & Parallel Processing Unit (T-Head) \\
    adaLN & Adaptive Layer Normalization & $\text{FPS}_{\text{agg}}$ & 7-Camera Aggregate Throughput & PSNR & Peak Signal-to-Noise Ratio \\
    ADE & Average Displacement Error & FT / FW & Front Tele / Front Wide & RL / RR / RT & Rear Left / Right / Tele \\
    FBD & Fr\'echet BEV Distance & FTD & Fr\'echet Trajectory Distance & RoPE & Rotary Position Embedding \\
    BEV & Bird's-Eye View & FVD & Fr\'echet Video Distance & RSI & Recursive Self-Improvement \\
    CFG & Classifier-Free Guidance & HDMap & High-Definition Map & SIL / HIL & Software / Hardware-in-the-Loop \\
    CL / CR & Cross Left / Cross Right & IDM & Intelligent Driver Model & SLAM & Simultaneous Localization \& Mapping \\
    DCU & Deep Computing Unit (Hygon) & KV & Key--Value Cache & SSIM & Structural Similarity Index \\
    DiT & Diffusion Transformer & LAK & Layout-Aware Kernel Score & Subj. q. & Subjective Quality (CLIP-IQA+) \\
    DMD & Distribution Matching Distillation & LoFTR & Detector-Free Feature Matching & VJP & Vector--Jacobian Product \\
    FID & Fr\'echet Inception Distance & LPIPS & Learned Perceptual Patch Sim. & VLM & Vision--Language Model \\
    \bottomrule
  \end{tabularx}
\end{table}

\begin{table}[bp]
  \centering
  \caption{\textbf{Mathematical notation and symbol definitions.}}
  \label{tab:notation}
  \footnotesize
  \begin{tabularx}{\linewidth}{@{}>{\RaggedRight}p{3.6cm}>{\RaggedRight}X@{}}
    \toprule
    \hd{Symbol} & \hd{Definition} \\
    \midrule
    $\mathcal{C},\, C=7,\, v\in\mathcal{C}$ & Calibrated surround camera rig, total camera count, and camera index (\cref{sec:formulation}) \\
    $t,\, k,\, q,\, L=2,\, N_s,\, D,\, d_k$ & Latent frame index, causal chunk index, query index, chunk length ($2$ latents $=8$ frames), spatial patch tokens per camera frame, DiT width ($2048$), and attention head dimension ($128$) \\
    $X_{t,v},\, z_{t,v}$ & Recorded RGB video frame ($H\times W\times 3$) and 16-channel causal VAE latent tensor ($16\times h\times w$) for camera $v$ \\
    $m_{t,v},\, M_t,\, z^{\text{layout}}_{t,v},\, y$ & Per-camera projected 3D HDMap/box layout frame, 7-view layout tuple, its 16-channel VAE latent, and VLM scene caption \\
    $\sigma\in[0,1],\, \sigma_{\text{write}}=0.13$ & Rectified-flow noise level (shared across all 7 cameras per chunk) and light-noise KV-memory commit level \\
    $\mathcal{E},\, \mathcal{N}(v),\, \mathcal{S}(v)$ & Physical rig adjacency graph ($7$ undirected / $14$ directed edges), neighbor set, and inclusive attention set $\{v\}\cup\mathcal{N}(v)$ (\cref{eq:xview}) \\
    $G_\phi,\, v_{\text{teacher}},\, v_{\text{fake}}$ & Distilled few-step causal generator, frozen Stage~I bidirectional score teacher, and tracking fake-score network (\cref{eq:dmd}) \\
    $W_{\text{base}},\, \Delta W_{\mathcal{W}},\, \Delta W_{\mathcal{D}}$ & Frozen Stage~III weight matrix and rank-32 writer / denoiser LoRA adapters (\cref{eq:dual}) \\
    $\mathcal{L}_{\text{LAK}},\, \mathcal{L}_{\text{ctrl}},\, \mathcal{P}_\psi,\, \mathcal{I},\, \tau_{v,c}$ & Layout-Aware Kernel Score, calibrated 5-class DINOv3 layout probe loss, frozen probe head, active view--class pairs, and VAE codec floor (\cref{eq:probe,eq:kernel}) \\
    $d_b,\, w_b,\, \sigma_b$ & Feature dimension, branch weight, and calibrated Gaussian kernel bandwidth of perceptual encoder $b$ (\cref{eq:kernel}) \\
    $\mathbf{R}(\cdot),\, T_{\max},\, W,\, s,\, o,\, \Delta$ & 1D temporal rotary operator, pre-tabulated RoPE horizon ($128$ latents), local KV window ($6$ latents), next global latent start index $s$, logical origin offset $o$, and exact rebase shift $\Delta=s-o-W$ (\cref{eq:rebase}) \\
    $z_{\text{src}},\, z_{\text{gt}},\, \sigma_{\text{src}}=0.3$ & Degraded 3DGS latent, paired ground-truth latent, and calibrated Fixer source-perturbation scale (\cref{eq:bridge}) \\
    \bottomrule
  \end{tabularx}
\end{table}

\section{Camera Rig, Architecture Hyperparameters, and System Details}\label{sec:config}\label{sec:system}
This section documents the calibrated seven-camera sensor geometry (\cref{tab:rig}), complete 480p and 720p architectural and Stage~IV training hyperparameters (\cref{tab:config}), steady-state 7-camera aggregate throughput benchmarks (\cref{tab:supp_efficiency}), and the closed-loop traffic dynamics formulation used by the interactive simulator.

\begin{table}[tbp]
  \centering
  \caption{\textbf{Seven-camera surround rig specification.} Extrinsic yaw and intrinsic horizontal field of view used across 3D layout projection, cross-view attention, and LoFTR evaluation.}
  \label{tab:rig}
  \footnotesize
  \begin{tabularx}{\linewidth}{@{}>{\RaggedRight\bfseries}p{3.0cm}cc>{\RaggedRight}X@{}}
    \toprule
    \hd{Camera} & \hd{HFOV} & \hd{Mount Yaw} & \hd{Primary Sensing Role} \\
    \midrule
    Front Wide (\texttt{FW}) & $120^\circ$ & $-0.3^\circ$ & Primary forward roadway, intersections, and ego lane geometry \\
    Front Tele (\texttt{FT}) & $30^\circ$ & $+0.1^\circ$ & Long-range lead vehicles, traffic lights, and highway horizon \\
    Cross Left (\texttt{CL}) & $120^\circ$ & $+49.9^\circ$ & Leftward cross traffic, unprotected turns, and crosswalks \\
    Cross Right (\texttt{CR}) & $120^\circ$ & $-49.8^\circ$ & Rightward merges, cut-ins, and curbside VRUs \\
    Rear Left (\texttt{RL}) & $70^\circ$ & $+129.9^\circ$ & Left adjacent-lane blind spot and overtaking vehicles \\
    Rear Right (\texttt{RR}) & $70^\circ$ & $-129.5^\circ$ & Right adjacent-lane blind spot and merge clearance \\
    Rear Tele (\texttt{RT}) & $30^\circ$ & $+180.0^\circ$ & Closing rear traffic directly behind ego vehicle \\
    \bottomrule
  \end{tabularx}
\end{table}

\begin{table}[tbp]
  \centering
  \caption{\textbf{Model and runtime configuration at 480p and 720p ($1280\times 704$).} All Stage~IV checkpoints share identical backbone weights and differ only in spatial token grid and role-specific LoRA adapters.}
  \label{tab:config}
  \scriptsize
  \setlength{\tabcolsep}{2.0pt}
  \begin{tabular*}{\linewidth}{@{\extracolsep{\fill}}lcc@{}}
    \toprule
    \hd{Parameter} & \hd{480p Real-Time Tier} & \hd{720p High-Fidelity Tier} \\
    \midrule
    Per-camera pixel resolution ($H\times W$) & $480\times 832$ & $704\times 1280$ \\
    VAE latent grid ($h\times w$, $4\times 8\times 8$ compression) & $60\times 104$ & $88\times 160$ \\
    Post-patchification tokens per camera latent ($N_s$) & $1{,}560$ & $3{,}520$ \\
    Tokens per 7-camera training clip (24 latents / 93 frames) & $262{,}080$ & $591{,}360$ \\
    Streaming chunk size (latent / display frames at 10\,Hz) & $2\;/\;8$ ($0.8$\,s) & $2\;/\;8$ ($0.8$\,s) \\
    Bounded local KV cache window (6 latents, $2.4$\,s) & $14.0$\,GiB total ($2.0$\,GiB/GPU) & $31.6$\,GiB total ($4.5$\,GiB/GPU) \\
    Stage~IV seeds per condition ($m$) / anchor weight ($\lambda_{\text{anc}}$) & $m=4$ / $\lambda_{\text{anc}}=0.30$ & $m=2$ / $\lambda_{\text{anc}}=0.05$ \\
    Stage~IV optimizer, peak LR, clip, population (32 GPUs) & AdamW, $10^{-5}$, clip $1.0$, pop.\ $128$ & AdamW, $10^{-5}$, clip $1.0$, pop.\ $128$ \\
    \midrule
    \multicolumn{3}{@{}l}{\textbf{Backbone:} 28 DiT blocks, width $D=2048$, 16 heads ($d_k=128$), 14-directed-edge gated cross-view attention (init.\ std.\ $10^{-3}$)} \\
    \multicolumn{3}{@{}l}{\textbf{Sampling:} Shifted flow timesteps $\{1000,750,500,250\}$ (subsets for $N_{\text{step}}\in\{1,2\}$), $\text{CFG}=1$, $\sigma_{\text{write}}=0.13$} \\
    \multicolumn{3}{@{}l}{\textbf{Stage IV:} Rank-32 $\Delta W_{\mathcal{W}},\Delta W_{\mathcal{D}}$ LoRA, 48-latent ($18.9$\,s) horizon + FIFO replay pool, memory VJP depth $H=2$, EMA $0.97$} \\
    \bottomrule
  \end{tabular*}
  \tablenote{Per-GPU KV memory divides the 7-camera total evenly across the 7 context-parallel camera ranks ($14.0/7=2.0$\,GiB at 480p; $31.6/7=4.51\approx 4.5$\,GiB at 720p).}
\end{table}

\paragraph{Stage~IV optimization and gradient verification.}
Across both resolution tiers, Stage~IV Antiphon Forcing freezes the Stage~III base backbone and updates only the decoupled writer ($\Delta W_{\mathcal{W}}$) and denoiser ($\Delta W_{\mathcal{D}}$) LoRA adapters with AdamW (peak learning rate $1\times 10^{-5}$, gradient-norm clip $1.0$, EMA decay $0.97$) over 48-latent-frame ($18.9$\,s) rollouts on 32 GPUs (population size $128$). The causal-VAE decoder-VJP memory backpropagation through depth $H=2$ preceding writer chunks was validated numerically against central finite differences on the 480p checkpoint: analytic and finite-difference gradients agree to $0.26\%$--$0.73\%$ relative error across writer value projections (blocks 25 and 27) and denoiser output projections, at $36.2$\,GiB peak per-GPU memory.

\paragraph{Reference profiling platform and aggregate throughput.}
All latency and frame-rate measurements (\cref{fig:latency,tab:supp_efficiency}) are profiled on an idle 8-GPU node with context parallelism assigning one camera per rank across seven GPUs, discarding 2 warm-up chunks and taking steady-state medians across 14 streaming chunks timed with CUDA events (including host orchestration and LightTAE decoding). While the main paper strictly reports synchronous \emph{per-camera FPS} ($8000/\text{Wall}_{\text{ms}}$) for direct comparison against the $10$\,Hz camera clock, \cref{tab:supp_efficiency} additionally provides the \textbf{7-camera aggregate image throughput} ($\text{FPS}_{\text{agg}}=56000/\text{Wall}_{\text{ms}}=7\times\text{FPS}$) to facilitate hardware-utilization comparisons across heterogeneous multi-stream engines.

\begin{table}[tbp]
  \centering
  \caption{\textbf{Supplementary efficiency table including 7-camera aggregate frame throughput ($\text{FPS}_{\text{agg}}$).} $\text{FPS}_{\text{agg}}=7\times\text{FPS}$ counts total rendered camera images per second across all seven streams ($56$ images per chunk), whereas synchronous simulation rate is governed by per-camera $\text{FPS}$.}
  \label{tab:supp_efficiency}
  \footnotesize
  \setlength{\tabcolsep}{4.5pt}
  \begin{tabular*}{\linewidth}{@{\extracolsep{\fill}}llcccc@{}}
    \toprule
    \hd{Resolution \& Sweep} & \hd{Configuration} & \hd{Wall (ms) $\downarrow$} & \hd{Per-Cam.\ FPS $\uparrow$} & \hd{Aggregate $\text{FPS}_{\text{agg}}$ (7 cams) $\uparrow$} & \hd{Speedup} \\
    \midrule
    480p (4-step) & Eager baseline & 1577 & 5.07 & 35.51 & $1.00\times$ \\
    480p (4-step) & FP8 attention & 1467 & 5.45 & 38.17 & $1.07\times$ \\
    480p (4-step) & \texttt{torch.compile} & 1438 & 5.56 & 38.94 & $1.10\times$ \\
    480p (4-step) & \texttt{compile} + FP8 & 1206 & 6.63 & 46.43 & $1.31\times$ \\
    \midrule
    480p (Step scaling) & 2-step \texttt{compile} + FP8 & 928 & 8.62 & 60.34 & $1.70\times$ \\
    \rowcolor{boschblue!7}\textbf{480p (Step scaling)} & \textbf{1-step \texttt{compile} + FP8} & \textbf{737} & \textbf{10.85} & \textbf{75.98} & $\mathbf{2.14\times}$ \\
    \midrule
    720p (4-step) & Eager baseline & 4500 & 1.78 & 12.44 & $1.00\times$ \\
    720p (4-step) & \texttt{torch.compile} & 4166 & 1.92 & 13.44 & $1.08\times$ \\
    720p (4-step) & FP8 attention & 3945 & 2.03 & 14.20 & $1.14\times$ \\
    \bottomrule
  \end{tabular*}
\end{table}

\paragraph{Closed-loop dynamics formulation.}\label{sec:agent_interface}
In closed-loop simulation (\cref{sec:closedloop,fig:closedloop}), external ego control commands $(a_{\text{ego}},\delta_{\text{ego}})$ are applied at each $800$\,ms chunk boundary ($1.25$\,Hz control update rate) and integrated internally at $10$\,Hz via a kinematic bicycle model~\citep{kong2015bicycle}, while interactive traffic agents follow the Intelligent Driver Model (IDM)~\citep{treiber2000idm}:
\begin{equation}
  \dot{v}_i = a_{\max}\!\left[1-\left(\frac{v_i}{v_0}\right)^4-\left(\frac{s^*(v_i,\Delta v_i)}{s_i}\right)^2\right],\qquad s^*(v_i,\Delta v_i)=s_0+\max\!\left(0,\,v_i T_{\text{gap}}+\frac{v_i\Delta v_i}{2\sqrt{a_{\max}b_{\text{comf}}}}\right),
\end{equation}
where $s_i$ and $\Delta v_i$ denote net bumper-to-bumper headway and closing speed relative to the counterfactual ego or lead vehicle. Updated 3D agent bounding boxes and static HDMap vectors are projected onto all seven cameras ahead of each $800$\,ms chunk.

\section{Evaluation Protocol, Trajectory Eligibility, and Full Per-View Tables}\label{sec:protocol}\label{sec:pervview}\label{sec:trajectory}
This section summarizes the adapted DrivingGen~\citep{drivinggen} and surround-view evaluation metrics (\cref{tab:protocol}), complete monocular-SLAM ego-trajectory results with valid scene counts (\cref{tab:trajectory}), and full per-camera breakdowns across all seven calibrated views (\cref{tab:pervview_all,tab:pervview_ablation}).

\paragraph{SLAM trajectory eligibility and resolution scaling.}
Monocular-SLAM ego-trajectory evaluation (\cref{tab:trajectory}) extracts camera motion from generated \texttt{FW} video via UniDepthV2~\citep{piccinelli2025unidepthv2}. Because SLAM tracking fails when generated footage exhibits severe structural warping or temporal flicker, valid scene count serves as a joint reliability indicator alongside ADE, DTW, and FTD. Unconditioned single-view baselines fail on $9.7\%$--$99.8\%$ of scenes (\emph{Wan2.2} succeeds on only $1/600$ scenes and is therefore omitted from population aggregates), whereas layout-grounded \emph{BoschDreams\textsuperscript{7V}} recovers $544/600$ ($90.7\%$) valid scenes at 480p 1-step ($7.02$\,m ADE, $411.56$ DTW, $65.39$ FTD), improves monotonically to $571/600$ ($95.2\%$, $5.17$\,m ADE) at 480p 4-step, and peaks at $573/600$ ($95.5\%$, $4.90$\,m ADE, $284.82$ DTW) at 720p 4-step.

\paragraph{Per-camera optical analysis.}
\Cref{tab:pervview_all,tab:pervview_ablation} disaggregate all seven cameras across primary and ablation checkpoints. Two optical effects govern per-camera variation: (i)~narrow-FOV telephoto views (\texttt{FT} $30^\circ$, \texttt{RT} $30^\circ$) amplify apparent pixel velocity and therefore exhibit slightly higher FVD ($86.00$--$97.99$) despite achieving the highest CLIP-IQA+ quality ($0.5706$--$0.5942$) and temporal consistency ($0.9073$--$0.9163$) at 720p; and (ii)~comparing Stage~IV LAK against full Antiphon Forcing at 480p 4-step confirms that LAK's lower 7V FVD mean ($80.97$ vs.\ $92.02$) arises strictly from side and rear views (\texttt{RL} $66.68\to 95.23$, \texttt{RR} $69.07\to 89.86$), whereas full Antiphon Forcing improves primary forward views (\texttt{FW} $95.02\to 83.59$, \texttt{FT} $113.01\to 101.12$) as well as tracked agent consistency and persistence across every camera.

\begin{table}[tbp]
  \centering
  \caption{\textbf{Summary of DrivingGen~\citep{drivinggen} and surround-view evaluation metrics adapted to our 600-scene benchmark.}}
  \label{tab:protocol}
  \scriptsize
  \setlength{\tabcolsep}{3.5pt}
  \begin{tabularx}{\linewidth}{@{}>{\bfseries}p{2.3cm}c>{\RaggedRight}X@{}}
    \toprule
    \hd{Metric} & \hd{Dir.} & \hd{Computation and Scope} \\
    \midrule
    Obj.\ q. & $\uparrow$ & Luminance FFT exposure/flicker stability across overlapping $3$\,s windows at $10$\,Hz ($600/600$ scenes per view). \\
    Subj.\ q. & $\uparrow$ & Learned perceptual frame quality via CLIP-IQA+~\citep{wang2023clipiqa} on $512\times 512$ resampled frames averaged over 100 frames. \\
    V-cons. & $\uparrow$ & Mean adjacent-frame DINOv3~\citep{oquab2025dinov3} cosine similarity on motion-selected keyframes via SEA-RAFT~\citep{wang2024searaft} optical-flow arc length. \\
    FVD & $\downarrow$ & 100-frame evaluation around the StyleGAN-V~\citep{skorokhodov2022styleganv} I3D Fr\'echet distance~\citep{unterthiner2018fvd} over 600-clip populations per camera. \\
    A-cons. & $\uparrow$ & Tracked-agent DINOv3~\citep{oquab2025dinov3} appearance similarity (initial + adjacent crops), averaged over valid tracks and scenes. \\
    A-pres. & $\uparrow$ & $1-\text{fraction}$ of tracked agents exhibiting unnatural disappearance as audited by Cosmos-Reason1-7B~\citep{nvidia2025cosmosreason1} VLM. \\
    CV Match / Conf. & $\uparrow$ & Outdoor LoFTR~\citep{sun2021loftr} correspondences ($\text{conf.}\ge 0.2$) and confidence sum across 7 adjacent camera pairs at 8 timestamps. \\
    ADE / DTW / FTD & $\downarrow$ & UniDepthV2~\citep{piccinelli2025unidepthv2} SLAM-aligned 101-step Euclidean error (m), dynamic time-warping cost, and Motion-Transformer~\citep{shi2022mtr} Fr\'echet distance. \\
    \bottomrule
  \end{tabularx}
\end{table}

\begin{table}[tbp]
  \centering
  \caption{\textbf{Complete ego-trajectory metrics and valid SLAM scene counts across all evaluated models.} Runs with fewer than four valid SLAM scenes are omitted (\na) from population trajectory aggregates (e.g., \emph{Wan2.2}, $1/600$ valid scene, $99.8\%$ SLAM failure rate). Bold marks the best and underline the second best among 600-scene models.}
  \label{tab:trajectory}
  \fontsize{8.5pt}{9pt}\selectfont
  \renewcommand{\arraystretch}{1.05}
  \setlength{\tabcolsep}{1.2pt}
  \begin{tabular}{@{}lc ccccc@{}}
    \toprule
    \hd{Method} & \hd{Valid / Req.} & \hd{ADE (m) $\downarrow$} & \hd{DTW $\downarrow$} & \hd{FTD $\downarrow$} & \hd{Traj.\ Quality $\uparrow$} & \hd{Traj.\ Consist.\ $\uparrow$} \\
    \midrule
    \rowcolor{boschgray!8}\textit{GT Measured Reference (annotations)} & 600 / 600 & \textit{0.00} & \textit{0.00} & \textit{0.00} & \textit{0.7469} & \textit{0.5676} (532) \\
    \midrule
    Causal Forcing~\citep{zhu2026causalforcing} & 124 / 600 & 20.70 & 1870.54 & 89.09 & 0.4277 & 0.3150 \\
    LingBot-World~\citep{team2026lingbot} & 318 / 600 & 16.96 & 1218.46 & 134.81 & 0.4083 & 0.3468 \\
    Wan2.2~\citep{wan2025,wan2025wan22} & 1 / 600 & \na & \na & \na & 0.4342 & 0.1484 \\
    FastVideo Causal Wan2.2~\citep{zhang2025fastvideo,zhang2025vsa} & 462 / 600 & 23.09 & 1619.89 & 88.89 & 0.4659 & 0.3571 \\
    HY-WorldPlay~\citep{sun2025worldplay} & 542 / 600 & 32.03 & 2762.43 & 86.63 & 0.3532 & 0.3005 \\
    Self Forcing~\citep{huang2025selfforcing} & 395 / 600 & 24.90 & 1967.60 & 153.05 & 0.4416 & 0.3462 \\
    Cosmos-Transfer~2.5 (First 100)~\citep{nvidia2025cosmos25} & 530 / 600 & 14.50 & 807.43 & 148.41 & 0.3962 & 0.3203 \\
    Cosmos-Transfer~2.5 (Last 100)~\citep{nvidia2025cosmos25} & 509 / 600 & 15.92 & 969.88 & 213.85 & 0.3907 & 0.3169 \\
    Cosmos~2.5~1V~\citep{nvidia2025cosmos25} & 477 / 600 & 20.13 & 1427.14 & 106.81 & 0.3711 & 0.3448 \\
    OmniDreams-1V$^\dagger$~\citep{nvidia2026omnidreams} & 490 / 600 & 13.49 & 791.57 & 80.42 & 0.4020 & 0.3642 \\
    \midrule
    (1) Stage~I Bidirectional Teacher (35-step, CFG 5) & 559 / 600 & 7.56 & 499.16 & 71.34 & 0.5059 & 0.4112 \\
    (2) Stage~II Causal Student (35-step, CFG 5) & 470 / 600 & 16.41 & 715.56 & 96.93 & 0.3806 & 0.3402 \\
    (3) Stage~III Self-Forcing (rollout DMD only) & 489 / 600 & 10.27 & 724.07 & 72.70 & 0.3979 & 0.3554 \\
    (4) + Stage~IV LAK (shared LoRA, no mem.\ VJP) & 561 / 600 & 5.39 & 367.13 & \secbest{50.10} & 0.4957 & 0.3864 \\
    BoschDreams\textsuperscript{7V} (480p, 1-step) & 544 / 600 & 7.02 & 411.56 & 65.39 & 0.5006 & 0.4199 \\
    BoschDreams\textsuperscript{7V} (480p, 2-step) & 552 / 600 & 5.66 & 353.07 & 57.94 & 0.5343 & 0.4207 \\
    BoschDreams\textsuperscript{7V} (480p, 4-step) & \secbest{571 / 600} & 5.17 & \secbest{300.74} & 56.92 & \secbest{0.5509} & 0.4305 \\
    BoschDreams\textsuperscript{7V} (720p, 1-step) & 556 / 600 & 6.10 & 392.43 & 60.17 & 0.4782 & \secbest{0.4311} \\
    BoschDreams\textsuperscript{7V} (720p, 2-step) & 564 / 600 & \secbest{5.15} & 327.08 & \best{49.18} & 0.5138 & \best{0.4377} \\
    \rowcolor{boschblue!7}\textbf{BoschDreams\textsuperscript{7V} (720p, 4-step)} & \best{573 / 600} & \best{4.90} & \best{284.82} & 55.48 & \best{0.5742} & 0.4281 \\
    \bottomrule
  \end{tabular}
  \tablenote{$^\dagger$In-house single-view HDMap-conditioned reimplementation following OmniDreams-SV~\citep{nvidia2026omnidreams}.}
\end{table}

\begin{table}[tbp]
  \centering
  \caption{\textbf{Per-view metrics for the 600-scene GT Measured Reference and primary BoschDreams\textsuperscript{7V} checkpoints.} GT-only FVD is not computed (\na). Conditional valid scene counts out of 600 are shown in parentheses for agent metrics; 7V means are computed from unrounded per-view values (e.g., $84.35$ for 720p 4-step).}
  \label{tab:pervview_all}
  \fontsize{8.5pt}{9.5pt}\selectfont
  \renewcommand{\arraystretch}{1.0}
  \setlength{\tabcolsep}{8.2pt}
   \begin{tabular}{@{}ll cccccc@{}}
    \toprule
    \hd{Checkpoint} & \hd{Camera View} & \hd{Obj.q.$\uparrow$} & \hd{Subj.q.$\uparrow$} & \hd{V-cons.$\uparrow$} & \hd{FVD$\downarrow$} & \hd{A-cons.\ (n)$\uparrow$} & \hd{A-pres.\ (n)$\uparrow$} \\
    \midrule
    \multirow{8}{2.8cm}{\raggedright \textit{GT Measured Reference (600)}}
      & Front Wide (\texttt{FW}) & 0.7633 & 0.5128 & 0.8874 & \na & 0.7956 (523) & 0.6088 (524) \\
      & Front Tele (\texttt{FT}) & 0.7817 & 0.5361 & 0.9242 & \na & 0.8128 (567) & 0.5626 (567) \\
      & Cross Left (\texttt{CL}) & 0.7567 & 0.5375 & 0.8531 & \na & 0.7481 (412) & 0.6731 (413) \\
      & Cross Right (\texttt{CR}) & 0.7317 & 0.5312 & 0.8899 & \na & 0.7755 (392) & 0.6684 (392) \\
      & Rear Tele (\texttt{RT}) & 0.7783 & 0.5802 & 0.9081 & \na & 0.8022 (532) & 0.6560 (532) \\
      & Rear Left (\texttt{RL}) & 0.7817 & 0.5463 & 0.8506 & \na & 0.7558 (408) & 0.7335 (409) \\
      & Rear Right (\texttt{RR}) & 0.7433 & 0.5374 & 0.8821 & \na & 0.7655 (402) & 0.7662 (402) \\
      & \textbf{7V Mean} & \textbf{0.7624} & \textbf{0.5402} & \textbf{0.8851} & \na & \textbf{0.7794} & \textbf{0.6669} \\
    \midrule
    \multirow{8}{2.8cm}{\raggedright \textbf{BoschDreams\textsuperscript{7V}} (480p, 1-step)}
      & Front Wide (\texttt{FW}) & 0.8567 & 0.4772 & 0.8531 & 74.93 & 0.7090 (521) & 0.6111 (522) \\
      & Front Tele (\texttt{FT}) & 0.9117 & 0.5008 & 0.8946 & 86.77 & 0.7074 (561) & 0.5704 (561) \\
      & Cross Left (\texttt{CL}) & 0.8200 & 0.4799 & 0.8248 & 78.04 & 0.6362 (405) & 0.6494 (405) \\
      & Cross Right (\texttt{CR}) & 0.7683 & 0.4797 & 0.8644 & \textbf{71.91} & 0.6656 (393) & 0.6336 (393) \\
      & Rear Tele (\texttt{RT}) & 0.9183 & 0.5286 & 0.8602 & 94.90 & 0.7116 (534) & 0.6330 (534) \\
      & Rear Left (\texttt{RL}) & 0.8400 & 0.4993 & 0.8057 & 76.89 & 0.6673 (400) & 0.7375 (400) \\
      & Rear Right (\texttt{RR}) & 0.8417 & 0.4903 & 0.8427 & 77.90 & 0.6769 (397) & 0.7186 (398) \\
      & \textbf{7V Mean} & \textbf{0.8510} & \textbf{0.4937} & \textbf{0.8494} & \textbf{80.19} & \textbf{0.6820} & \textbf{0.6505} \\
    \midrule
    \multirow{8}{2.8cm}{\raggedright \textbf{BoschDreams\textsuperscript{7V}} (720p, 4-step)}
      & Front Wide (\texttt{FW}) & 0.8834 & 0.5656 & 0.8962 & 83.12 & 0.7594 (523) & 0.6654 (524) \\
      & Front Tele (\texttt{FT}) & 0.9103 & 0.5706 & 0.9163 & 86.00 & 0.7502 (561) & 0.5804 (562) \\
      & Cross Left (\texttt{CL}) & 0.8902 & 0.5516 & 0.8710 & 87.20 & 0.7024 (412) & 0.6993 (413) \\
      & Cross Right (\texttt{CR}) & 0.8913 & 0.5463 & 0.8912 & \textbf{73.80} & 0.7300 (385) & 0.6637 (385) \\
      & Rear Tele (\texttt{RT}) & 0.9290 & 0.5942 & 0.9073 & 97.99 & 0.7547 (530) & 0.6712 (531) \\
      & Rear Left (\texttt{RL}) & 0.9010 & 0.5639 & 0.8534 & 81.81 & 0.7295 (406) & 0.7961 (406) \\
      & Rear Right (\texttt{RR}) & 0.8952 & 0.5594 & 0.8907 & 80.49 & 0.7254 (398) & 0.7450 (400) \\
      & \textbf{7V Mean} & \textbf{0.9001} & \textbf{0.5645} & \textbf{0.8894} & \textbf{84.35} & \textbf{0.7359} & \textbf{0.6887} \\
    \bottomrule
  \end{tabular}
\end{table}

\begin{table}[tbp]
  \centering
  \caption{\textbf{Per-view ablation checkpoints (480p 2-step, 480p 4-step, Stage IV LAK, and Stage III Self-Forcing).}}
  \label{tab:pervview_ablation}
  \fontsize{8.0pt}{9.0pt}\selectfont
  \renewcommand{\arraystretch}{1.0}
  \setlength{\tabcolsep}{2.2pt}
  \begin{tabular*}{\linewidth}{@{\extracolsep{\fill}}ll cccccc@{}}
    \toprule
    \hd{Model / Checkpoint} & \hd{Camera View} & \hd{Obj.q.$\uparrow$} & \hd{Subj.q.$\uparrow$} & \hd{V-cons.$\uparrow$} & \hd{FVD$\downarrow$} & \hd{A-cons.\ (n)$\uparrow$} & \hd{A-pres.\ (n)$\uparrow$} \\
    \midrule
    \multirow{8}{*}{BoschDreams\textsuperscript{7V} (480p, 2-step)}
      & Front Wide (\texttt{FW}) & 0.8733 & 0.4770 & 0.8571 & 78.12 & 0.7137 (522) & 0.6508 (524) \\
      & Front Tele (\texttt{FT}) & 0.9267 & 0.5017 & 0.8965 & 87.51 & 0.7102 (562) & 0.5890 (562) \\
      & Cross Left (\texttt{CL}) & 0.7967 & 0.4809 & 0.8280 & 82.55 & 0.6410 (405) & 0.6247 (405) \\
      & Cross Right (\texttt{CR}) & 0.8067 & 0.4788 & 0.8654 & 74.77 & 0.6703 (393) & 0.6447 (394) \\
      & Rear Tele (\texttt{RT}) & 0.9417 & 0.5274 & 0.8642 & 90.50 & 0.7132 (534) & 0.6629 (534) \\
      & Rear Left (\texttt{RL}) & 0.8533 & 0.4994 & 0.8053 & 76.02 & 0.6721 (403) & 0.7543 (403) \\
      & Rear Right (\texttt{RR}) & 0.8633 & 0.4909 & 0.8498 & 77.03 & 0.6808 (393) & 0.7538 (394) \\
      & \textbf{7V Mean} & \textbf{0.8660} & \textbf{0.4937} & \textbf{0.8523} & \textbf{80.93} & \textbf{0.6859} & \textbf{0.6686} \\
    \midrule
    \multirow{8}{*}{BoschDreams\textsuperscript{7V} (480p, 4-step)}
      & Front Wide (\texttt{FW}) & 0.8400 & 0.4714 & 0.8622 & 83.59 & 0.7215 (523) & 0.6343 (525) \\
      & Front Tele (\texttt{FT}) & 0.9250 & 0.5002 & 0.9029 & 101.12 & 0.7217 (560) & 0.5768 (560) \\
      & Cross Left (\texttt{CL}) & 0.8583 & 0.4727 & 0.8277 & 91.68 & 0.6516 (407) & 0.6486 (407) \\
      & Cross Right (\texttt{CR}) & 0.8450 & 0.4751 & 0.8688 & 87.23 & 0.6781 (387) & 0.6589 (387) \\
      & Rear Tele (\texttt{RT}) & 0.8550 & 0.5235 & 0.8741 & 95.44 & 0.7210 (535) & 0.6374 (535) \\
      & Rear Left (\texttt{RL}) & 0.8617 & 0.4947 & 0.8110 & 95.23 & 0.6778 (401) & 0.7631 (401) \\
      & Rear Right (\texttt{RR}) & 0.8883 & 0.4868 & 0.8539 & 89.86 & 0.6892 (397) & 0.7254 (397) \\
      & \textbf{7V Mean} & \textbf{0.8676} & \textbf{0.4892} & \textbf{0.8572} & \textbf{92.02} & \textbf{0.6944} & \textbf{0.6635} \\
    \midrule
    \multirow{8}{*}{Stage IV LAK (shared LoRA)}
      & Front Wide (\texttt{FW}) & 0.9217 & 0.4773 & 0.8552 & 95.02 & 0.7119 (524) & 0.6008 (526) \\
      & Front Tele (\texttt{FT}) & 0.9667 & 0.4984 & 0.8950 & 113.01 & 0.7098 (559) & 0.5689 (559) \\
      & Cross Left (\texttt{CL}) & 0.8783 & 0.4816 & 0.8297 & 77.69 & 0.6487 (401) & 0.6576 (403) \\
      & Cross Right (\texttt{CR}) & 0.9000 & 0.4818 & 0.8671 & 69.95 & 0.6735 (389) & 0.6452 (389) \\
      & Rear Tele (\texttt{RT}) & 0.9367 & 0.5214 & 0.8689 & 75.35 & 0.7153 (533) & 0.6454 (533) \\
      & Rear Left (\texttt{RL}) & 0.9067 & 0.5028 & 0.8200 & 66.68 & 0.6735 (395) & 0.7096 (396) \\
      & Rear Right (\texttt{RR}) & 0.9117 & 0.4951 & 0.8552 & 69.07 & 0.6874 (391) & 0.7303 (393) \\
      & \textbf{7V Mean} & \textbf{0.9174} & \textbf{0.4941} & \textbf{0.8559} & \textbf{80.97} & \textbf{0.6886} & \textbf{0.6511} \\
    \midrule
    \multirow{8}{*}{Stage III Self-Forcing}
      & Front Wide (\texttt{FW}) & 0.8917 & 0.4452 & 0.8285 & 138.75 & 0.6931 (525) & 0.5856 (526) \\
      & Front Tele (\texttt{FT}) & 0.9483 & 0.4694 & 0.8800 & 143.79 & 0.6885 (562) & 0.5302 (562) \\
      & Cross Left (\texttt{CL}) & 0.8383 & 0.4568 & 0.8292 & 115.31 & 0.6296 (401) & 0.6434 (401) \\
      & Cross Right (\texttt{CR}) & 0.8900 & 0.4627 & 0.8561 & 102.97 & 0.6566 (391) & 0.5934 (391) \\
      & Rear Tele (\texttt{RT}) & 0.9217 & 0.5094 & 0.8399 & 154.62 & 0.6970 (532) & 0.6417 (533) \\
      & Rear Left (\texttt{RL}) & 0.8683 & 0.4837 & 0.8150 & 119.44 & 0.6573 (395) & 0.7013 (395) \\
      & Rear Right (\texttt{RR}) & 0.8983 & 0.4795 & 0.8367 & 124.23 & 0.6701 (390) & 0.7008 (391) \\
      & \textbf{7V Mean} & \textbf{0.8938} & \textbf{0.4724} & \textbf{0.8408} & \textbf{128.44} & \textbf{0.6703} & \textbf{0.6280} \\
    \bottomrule
  \end{tabular*}
\end{table}

\end{document}